\documentclass[journal]{IEEEtran}
\usepackage{amsmath,amsfonts}
\usepackage{algorithmic}
\usepackage{algorithm}
\usepackage{array}
\usepackage[caption=false,font=normalsize,labelfont=sf,textfont=sf]{subfig}
\usepackage{textcomp}
\usepackage{stfloats}
\usepackage{url}
\usepackage{verbatim}
\usepackage{graphicx}
\usepackage{cite}
\usepackage{balance}
\usepackage{multirow}
\usepackage{booktabs} 
\usepackage{color}
\newcommand{\red}[1]{{\textcolor{red}{ #1}}}
\newcommand{\blue}[1]{{\textcolor{blue}{ #1}}}

\newcommand{\etal} {\textit{et al.}}

\begin{document}

\title{OccCasNet: Occlusion-aware Cascade Cost Volume for Light Field Depth Estimation}

\author{Wentao Chao, Fuqing Duan, Xuechun Wang, Yingqian Wang, Guanghui Wang, ~\IEEEmembership{Senior Member,~IEEE}
\thanks{W. Chao, F. Duan, and X. Wang are with the School of
Artificial Intelligence, Beijing Normal University, Beijing 100875, China. (e-mail: chaowentao@mail.bnu.edu.cn; fqduan@bnu.edu.cn ; wangxuechun@mail.bnu.edu.cn).}
\thanks{Y. Wang is with the College of Electronic Science and Technology, National University of Defense Technology, Changsha 410073, China. (e-mail: wangyingqian16@nudt.edu.cn).}
\thanks{G. Wang is with the Department of Computer Science, Toronto Metropolitan University, Toronto, ON M5B 2K3, Canada. (e-mail: wangcs@torontomu.ca).}
\thanks{Manuscript received May 28, 2023. Corresponding author: F. Duan}}



\maketitle

\begin{abstract}

Light field (LF) depth estimation is a crucial task with numerous practical applications. 
However, mainstream methods based on the multi-view stereo (MVS) are resource-intensive and time-consuming as they need to construct a finer cost volume. 
To address this issue and achieve a better trade-off between accuracy and efficiency, we propose an occlusion-aware cascade cost volume for LF depth (disparity) estimation. 
Our cascaded strategy reduces the sampling number while keeping the sampling interval constant during the construction of a finer cost volume. We also introduce occlusion maps to enhance accuracy in constructing the occlusion-aware cost volume.
Specifically, we first obtain the coarse disparity map through the coarse disparity estimation network. 
Then, the sub-aperture images (SAIs) of side views are warped to the center view based on the initial disparity map. 
Next, we propose photo-consistency constraints between the warped SAIs and the center SAI to generate occlusion maps for each SAI. 
Finally, we introduce the coarse disparity map and occlusion maps to construct an occlusion-aware refined cost volume, enabling the refined disparity estimation network to yield a more precise disparity map.
Extensive experiments demonstrate the effectiveness of our method. Compared with state-of-the-art methods, our method achieves a superior balance between accuracy and efficiency and ranks first in terms of MSE and Q25 metrics among published methods on the HCI 4D benchmark. The code and model of the proposed method are available at \url{https://github.com/chaowentao/OccCasNet}.

\end{abstract}


\section{Introduction}
\label{sec:introduction}

\IEEEPARstart{L}{ight} field (LF)\cite{zhang2016light, chen2017light, zhang2019depth, cheng2019light, sheng2022urbanlf} images can simultaneously record the 4D spatial and angular information of light via a single snapshot. The 4D information can provide abundant cues, which is crucially needed in many practical application areas, such as refocusing  \cite{ng2005light, wang2018selective}, super-resolution \cite{zhang2019residual, jin2020light, cheng2021light, cheng2022spatial, chen2022light, van2023light, wang2023ntire}, view synthesis  \cite{wu2018light, meng2019high, jin2020deep}, semantic segmentation \cite{sheng2022urbanlf}, 3D reconstruction \cite{kim2013scene}, virtual reality \cite{yu2017light}, especially depth estimation \cite{zhang2016light, chen2017light, zhang2019depth, shin2018epinet,  tsai2020attention, peng2020zero, chen2021attention, wang2022occlusion, wang2022disentangling, chao2022learning}. By analyzing the properties of the LF images, we can accurately infer the depth of the scene. 

Currently, several methods have been proposed for LF depth estimation and achieve significant progress. These methods can be classified as traditional methods and learning-based methods. Traditional depth estimation algorithms \cite{tao2013depth, jeon2015accurate, williem2016robust, zhang2016light, chen2017light, zhu2017occlusion, zhang2019depth} fall under multi-view stereo (MVS) matching, Epipolar Plane Image (EPI), and defocus methods. However, these methods rely on hand-crafted features and a set of prior assumptions, which makes them computationally intensive and time-consuming for inference.

\begin{figure}[tb]
\centering
\includegraphics[width=\linewidth]{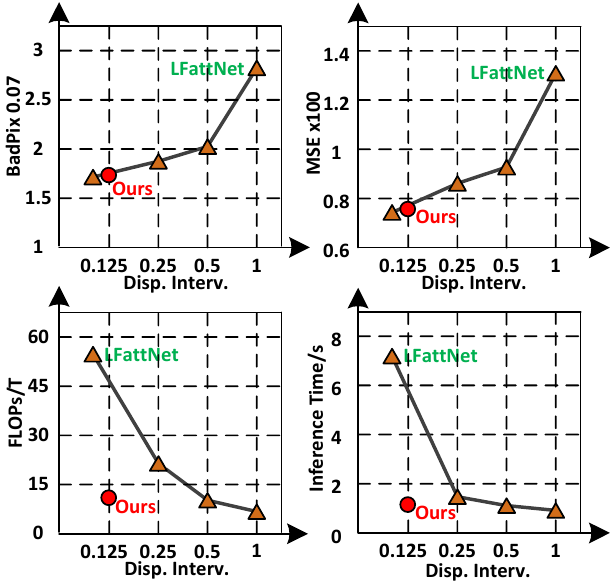}
\caption{The performance of the LFattNet \cite{tsai2020attention} was evaluated for different disparity intervals, with a predefined disparity range from -4 to 4. Badpix 0.07 and MSE $\times$100 report the average metrics in the HCI 4D benchmark \cite{honauer2016dataset} validation set. FLOPs and Inference Time are the average results of running LFattNet 10 times with an input LF image of 9$\times$9$\times$256$\times$256. Lower is better.}
\label{fig: intro}
\end{figure}

In recent years, deep learning has been rapidly developed and increasingly used for LF disparity estimation \cite{heber2016convolutional, he2018learning, shin2018epinet,  tsai2020attention, chen2021attention, wang2022occlusion, wang2022disentangling, chao2022learning}. 
Deep learning-based methods can leverage a vast amount of training data to extract prior knowledge of object features, instead of relying on feature analysis under various assumptions in traditional methods. 
Moreover, these methods can directly obtain scene depth after training, without requiring any post-processing operation. Compared with traditional techniques, deep learning-based methods have significantly improved accuracy and efficiency. 
Currently, the mainstream deep learning-based methods \cite{tsai2020attention, chen2021attention, wang2022occlusion, chao2022learning} are built on the theory of MVS matching, which involves four primary steps: feature extraction, cost volume construction, cost volume aggregation, and disparity regression. 
During the construction of the cost volume, the disparity interval and disparity number are hyper-parameters and set by human experience.
In this study, we employ LFattNet \cite{tsai2020attention} to explore the impact of different disparity intervals and sampling numbers on accuracy within a predefined disparity range.
As demonstrated in Fig.~\ref{fig: intro}, a finer disparity interval and an increased disparity number tend to improve the disparity accuracy. However, decreasing the cost volume disparity interval substantially increases FLOPs and inference time, limiting its practical application. 
In summary, two main challenges need to be addressed urgently for improving the speed and maintaining high accuracy in LF depth estimation. First, Fig.~\ref{fig: intro} confirms that constructing a finer cost volume significantly enhances the accuracy of LF depth estimation. 
However, the finer cost volume is constructed by \textit{shift-and-concat } \cite{tsai2020attention} operation in predefined disparity samplings, which is time-consuming and leads to high FLOPs during subsequent cost aggregation. 
Therefore, the challenge is to maintain the finer disparity interval while reducing the number of samples within the given disparity range. 
Second,  the cost volume does not consider occlusion during construction, which will mislead the matching of the cost volume in the subsequent aggregation process, resulting in a decrease in accuracy.
Existing methods \cite{tsai2020attention, chen2021attention, chao2022learning} treat pixels at different spatial locations equally during cost volume construction, which is not capable of managing spatially-varying occlusions where some views provide less informative data and can potentially impair the estimation results. OACC-Net \cite{wang2022occlusion} has addressed this issue by constructing an occlusion-aware cost volume at the pixel level using dilated convolution. However, sub-pixel cost volume construction is not supported due to integer dilated rate limitations. Therefore, the challenge is to develop an occlusion-aware cost volume at sub-pixel level construction method to further improve the accuracy.    

To address the aforementioned challenges, we propose an occlusion-aware cascade cost volume to estimate the disparity of LF images. 
First, we propose a coarse-to-fine approach to construct a cascade sub-pixel cost volume, consisting of coarse and refined levels. The coarse cost volume is constructed using a larger predefined disparity interval to cover the entire disparity range and produce a coarse disparity map. The refined cost volume at the sub-pixel level uses a smaller interval based on the coarse disparity map, and the disparity search range is adaptively adjusted in the refined stage.
Second, to handle occlusion situations in LF, we obtain occlusion maps for each view by leveraging the photo-consistency constraint based on the coarse disparity map. During the construction of the refined cost volume, we introduce occlusion maps to represent the importance of pixels from different views, which helps to alleviate the impact of occluded pixels and construct an occlusion-aware refined cost volume.

In summary, the contributions of our paper are as follows:
 \begin{itemize}
 
 \item We study the effect of disparity interval and disparity number in cost volume on accuracy and speed, and propose a method (named OccCasNet) for LF depth estimation. 

\item The cascade sub-pixel cost volume is constructed in a coarse-to-fine manner, which can save running time and maintain finer sub-pixel level disparity intervals. We introduce occlusion maps to construct an occlusion-aware refined cost volume and can alleviate occlusions by adaptively adjusting the weights of different views.  

\item Extensive experiments using both real camera images and synthetic LF images validate the effectiveness of our method. In comparison with state-of-the-art methods, our method achieves a superior trade-off between precision and speed and ranks first in terms of MSE and Q25 metrics on the HCI 4D benchmark.

\end{itemize}

\section{Related Work}
\label{sec:related}


In this section, we first introduce the LF representation, followed by a review of some traditional methods and deep learning-based methods for LF depth estimation. Finally, we discuss the issue of occlusion handling for LF images.


\begin{figure}[tb]
\centering
\includegraphics[width=\linewidth]{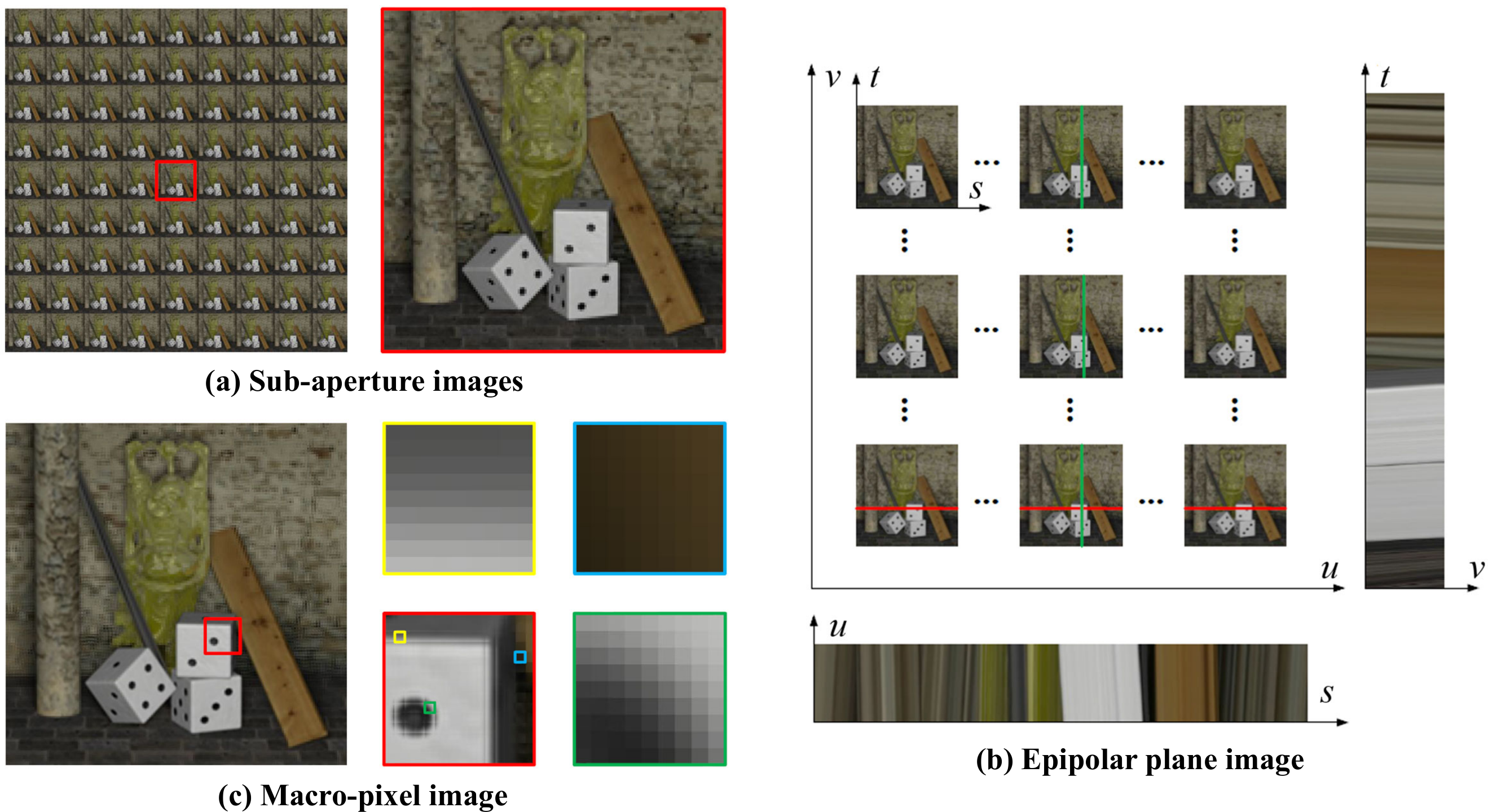}
\caption{Light field visualization. (a) Sub-aperture images (SAI). (b) Epipolar plane image (EPI). (c) Macro-pixel image (MacPI).}
\label{fig: related}
\end{figure}

\subsection{Light Field Representation}
In this paper, we uniformly use the two-plane model from Levoy\cite{levoy1996light} method to represent the LF image, $L(u,v,x,y)$, where $(u,v)$ represents the angular resolution, and $(x,y)$ represents the spatial resolution. When visualization of LF images, we also use $L(u,v,s,t)$. Both are equivalent, just for convenience. 
To visualize the structure of a 4D LF, we fix two variables in the 4D LF representation $L(u,v,x,y)$ and perform LF visualization in 2D space.
The first form is the sub-aperture images (SAI). By fixing $u = u^{*}$ and $v = v^{*}$, we can get a certain SAI $L(u^{*},v^{*},s,t)$ of multiple perspectives, as shown in Fig.~\ref{fig: related}(a). The second form is the epipolar plane image (EPI). We can fix $v = v^{*}$ and $t = t^{*}$ to obtain the horizontal EPI $L(u,v^{*},s,t^{*})$, as shown in the horizontal direction in Fig.~\ref{fig: related}(b). Similarly, we can get the vertical EPI $L(u^{*},v,s^{*},t)$ by fixing $u = u^{*}$ and $s = s^{*}$, as shown in the vertical direction in Fig.~\ref{fig: related}(b). The third form is the macro-pixel image (MacPI). By fixing $s = s^{*}$ and $t = t^{*}$, a specific macro-pixel $L(u,v,s^{*},t^{*})$ can be obtained, as shown in Fig.~\ref{fig: related}(c).

\subsection{Light Field Depth Estimation}
\subsubsection{Traditional Methods}
The MVS-based methods \cite{jeon2015accurate} obtain depth by exploiting the multi-view information of the SAIs for stereo matching. Jeon \etal \cite{jeon2015accurate} used phase translation theory to represent the sub-pixel translation between sub-aperture images. The center sub-aperture image was matched with other sub-aperture images to perform stereo matching. 
The EPI-based methods \cite{zhang2016robust} are able to reflect the depth of the scene by calculating the slope of the diagonal line in the EPI. Wanner \etal \cite{wanner2014variational} proposed a structure tensor to estimate the slope of lines in horizontal and vertical EPIs, and refined the initial results by global optimization. Zhang \etal \cite{zhang2016robust} proposed a Spinning Parallelogram Operator (SPO) to compute the slope of straight lines in EPI, which is insensitive to occlusion, noise, and spatial blending. 
The defocus-based method \cite{tao2013depth} measures the blur of a pixel at different focal stacks to obtain its corresponding depth. Williem \etal \cite{williem2016robust} used the information entropy between different angles and adaptive scattering to improve robustness to occlusion and noise. Tao \etal \cite{tao2013depth} combined scattering and matching cues to obtain a local depth map using Markov random field for global optimization. Zhang \etal \cite{zhang2016light} exploited the special linear structure of an epipolar plane image (EPI) and locally linear embedding (LLE) for LF depth estimation. Chen \etal \cite{chen2017light} proposed a sub-aperture scan and normalized fluctuation to acquire/calculate the scene disparity. Zhang \etal \cite{zhang2019depth} exploited graph-based structure-aware analysis and proposed a two-stage method for LF depth estimation. However, these methods rely on hand-designed features and subsequent optimization, which are time-consuming and have limited accuracy.

\subsubsection{Deep Learning-based Methods}
 
In recent years, deep learning has been rapidly developed and  widely applied in various LF processing tasks, especially depth estimation. Heber \etal \cite{heber2016convolutional} first used CNN to extract features from EPI and calculate the scene's depth. Shin \etal \cite{shin2018epinet} proposed EPINet with four directional (0°, 90°, 45°, and 135°) SAIs being set as input and the center viewpoint disparity map as output. Tsai \etal \cite{tsai2020attention} proposed the LFattNet network where a view selection module based on the attention mechanism is used to calculate the importance of each view and also serves as the weight for each view cost aggregation. Chen \etal \cite{chen2021attention} designed the AttMLFNet, an attention-based multilevel fusion network, including intra-branch and inter-branch fusion strategies, to fuse the features from different perspectives fusion hierarchically. Wang \etal \cite{wang2022disentangling} generalized the spatial-angular interaction mechanism to the disentangling mechanism and proposed DistgSSR for LF depth estimation.
Chao \etal \cite{chao2022learning} proposed the SubFocal method to learn the sub-pixel disparity distribution by constructing a sub-pixel cost volume and leveraging disparity distribution constraint, further obtaining a high-precision disparity map. However, existing algorithms have achieved high accuracy in LF depth estimation, they are time-consuming and do not obtain a good trade-off between accuracy and speed.


\subsection{Light Field Image with Occlusion}

Due to the dense sampling of angular views in LF images, occlusion has become a crucial issue in many LF applications, particularly depth estimation. In the case of Lambertian scenes, it is commonly assumed that pixels exhibit photo-consistency, which means that when focused to their depth, all viewpoints will converge to a single point. However, this assumption is no longer valid when occlusions are present. Wang \etal \cite{wang2015occlusion, wang2016depth} proposed that although the pixels in the occlusion may not exhibit photo-consistency, most of the viewing angles remain consistent. Additionally, the line that separates the occluded object from the occluder has the same orientation as the occlusion edge in the spatial domain. By ensuring photo-consistency in only the unoccluded view region, the accuracy of depth estimation can be improved. However, the algorithm's performance is affected by the accuracy of the angular patch division, and it is not suitable for handling complex occlusion situations where the occlusion cannot be divided linearly.
Williem \etal \cite{williem2016robust, williem2018robust} proposed a novel data cost that uses an angular entropy metric and an adaptive defocus response to enhance the algorithm's robustness against occlusions and noise. However, this approach can cause some regions to be over-smoothed, especially for complex details, due to the high randomness of the angular entropy metric.
Zhu \etal \cite{zhu2017occlusion} proposed an occluder-consistency approach that considers both spatial and angular domains, which can guide the selection of unoccluded views. They also designed an anti-occlusion energy function to optimize the depth map. However, the algorithm relies on the K-means clustering strategy, and it cannot handle situations with more complex clusters.
Chen \etal \cite{chen2018accurate} proposed a method to detect partially occluded boundary regions (POBR) using superpixel-based regularization and to handle occlusions from a POBR-based post-optimization perspective. However, this approach depends on the use of superpixels and requires multiple refinement steps to produce the final depth estimate.
Zhang \etal \cite{zhang2020depth} did not employ partial corner blocks for depth estimation. Instead, they used an undirected graph to jointly consider the occluded and unoccluded sub-aperture images (SAIs) in the corner blocks to exploit the structural information of the LF.
Han \etal \cite{han2022novel} introduced a novel approach for depth estimation that does not rely on photo-consistency as the primary metric for determining the correct depth. Instead, they proposed an occlusion-aware vote cost (OAVC) to preserve edges in the depth map more accurately. This method counts the number of refocused pixels that differ from the center view pixel by less than a small threshold and uses this count to select the correct depth. 
Guo \etal \cite{guo2020accurate} proposed an occlusion region detection network (ORDNet) for explicit estimation of occlusion maps, a coarse depth estimation network (CDENet), and a refined depth estimation network (RDENet), focusing on depth estimation of non-occluded and occluded regions, respectively, guided by the obtained occlusion maps. 
Wang \etal \cite{wang2022occlusion} proposed the OACC-Net, which was designed to build an occlusion-aware cost volume using dilated convolution without a disparity shift operation and iterative processing cost volume with occlusion masks.

The main distinctions between our method and the methods mentioned above are twofold. First, we construct a cascade cost volume that can achieve sub-pixel accuracy with fewer disparity numbers as compared to previous methods. Second, we explicitly generate occlusion maps for each view through a coarse disparity map based on photo-consistency constrain, which can effectively guide the construction of our refined cost volume and alleviate the impact of occluded views.

\section{Method}
\label{sec:method}

\begin{figure*}[tb]
  \centering
  \includegraphics[width=\linewidth]{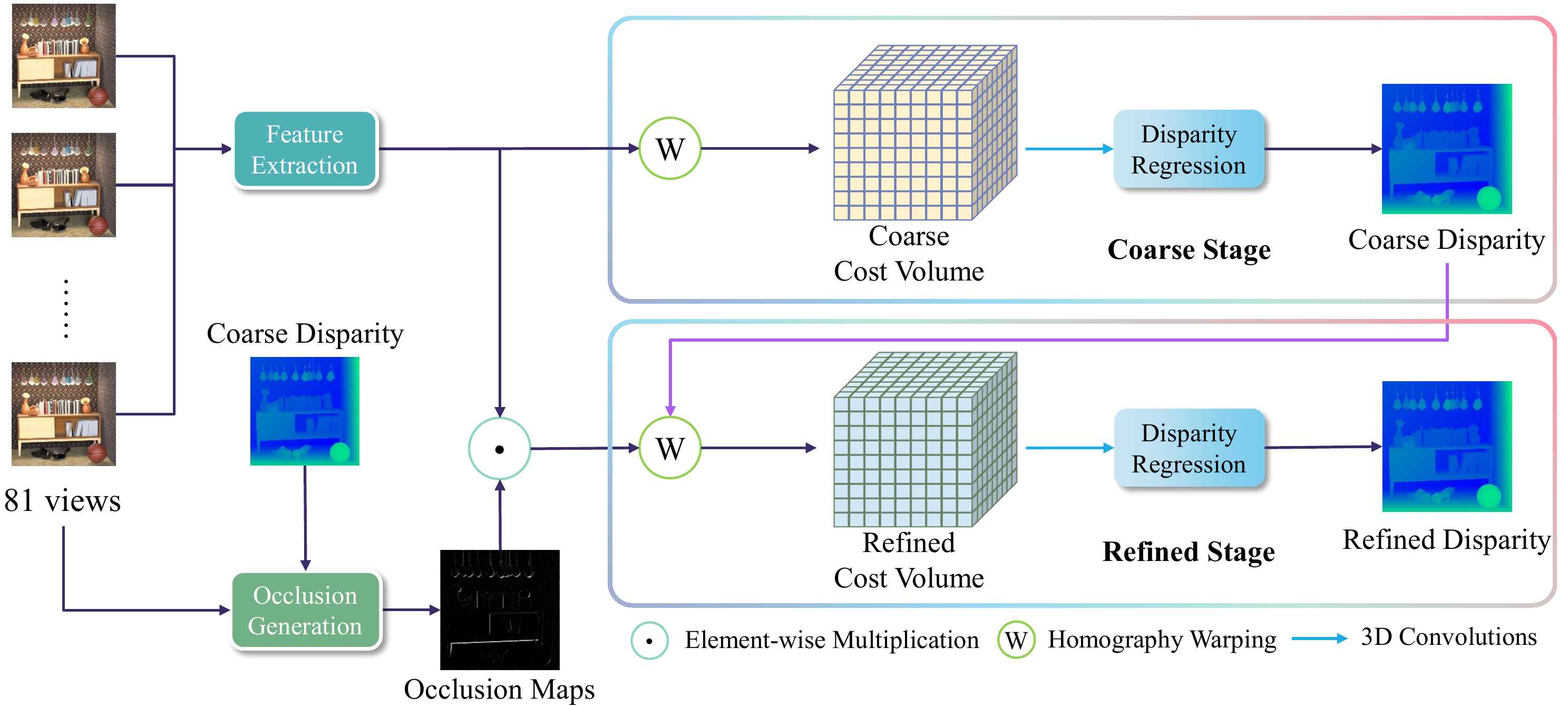}

  \caption{The overall architecture of the proposed OccCasNet. 
  The feature extraction module is utilized to extract the features of each SAI and form the shared feature map. The coarse disparity estimation network is adopted to generate the coarse disparity map. Additionally, the occlusion generation module is used to calculate the occlusion maps. Finally, the refined disparity estimation network takes both the coarse disparity and occlusion maps as inputs to generate a refined disparity map.}
  \label{fig: network}
\end{figure*}

In this section, we will explain how to construct occlusion-aware cascade cost volume. Then, we will provide an overview of the pipeline of OccCasNet and the loss function. Figure \ref{fig: network} depicts the overall framework of OccCasNet, which comprises feature extraction, coarse stage, occlusion maps generation, and refined stage. 

\begin{figure}[tb]
  \centering
  \includegraphics[width=\linewidth]{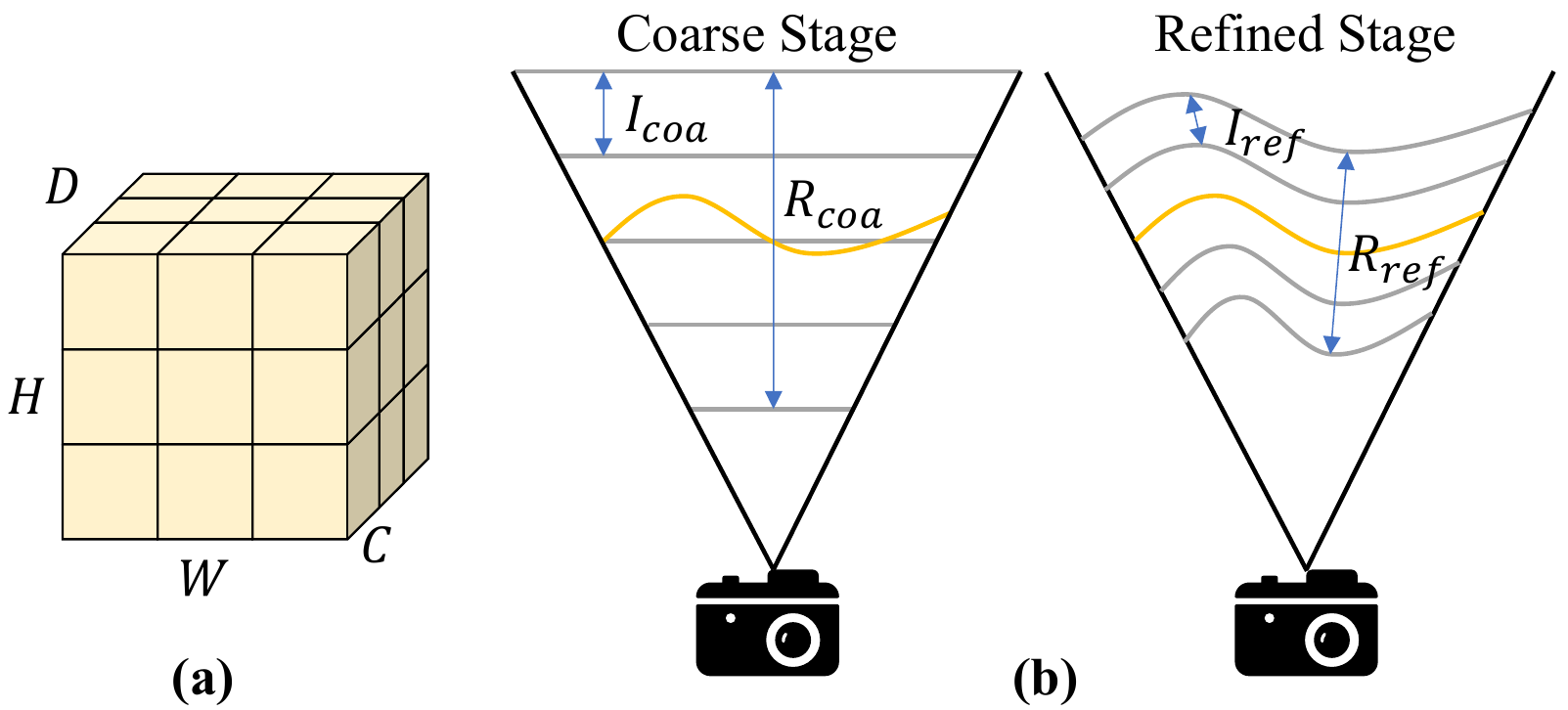}

  \caption{Illustration of the construction of cost volume. (a) A standard cost volume. $D$ is the disparity number, $H\times W$ denotes the spatial resolution, and $C$ is the channel number of feature map. (b) Illustration of disparity range of different stages. $R_{coa}$ and $I_{coa}$ respectively represent the disparity range and the disparity number for the coarse stage, while $D_{ref}$ and $I_{ref}$ respectively represent the disparity range and the disparity number for the refined stage. The gray lines in (b) denote the disparity range, and the yellow line indicates the predicted disparity map obtained from the coarse stage, which is used to determine the disparity range and disparity intervals for the refined stage.}
  \label{fig: cost}
\end{figure}

\subsection{Occlusion-aware Cascade Cost Volume}

\subsubsection{Cascade Cost Volume}

Figure \ref{fig: cost} (a) shows the standard cost volume of size $D \times H \times W \times C$, where $H \times W$ denotes the spatial resolution, $D$ is the disparity number, and $C$ is the channel number of feature maps. As analyzed above, an increased disparity number $D$ and a finer disparity interval are likely to improve the disparity accuracy. However, this will lead to a significant increase in computation and inference time. To resolve the problems, we propose a cascade cost volume and predict the output in a coarse-to-fine manner.

 \noindent \textbf{Disparity Interval and Disparity Range.} As depicted in Fig.~\ref{fig: cost} (b), $I_{coa}$ and $D_{coa}$ represent  the disparity interval and disparity range of the coarse stage, $I_{ref}$ and $D_{ref}$ represent the disparity interval and disparity range of the refined stage. 
 In the coarse stage, we can set a coarse disparity range $D_{coa}$ that covers the entire disparity range of the input scene with a larger coarse disparity interval $I_{coa}$. In the refined stages, we can narrow down the disparity range using a finer disparity interval based on the predicted disparity from the coarse stage. Consequently, the $I_{ref}$ and $D_{ref}$ can be set as: 

 \begin{equation}
\label{eq: cost}
\begin{aligned}
D_{ref} = D_{coa} * w_D, I_{ref} = I_{coa} * w_I,
\end{aligned}
\end{equation}
 \noindent where  $w_D < 1$ and $w_I < 1$ represent the decay factor of disparity range and disparity interval, respectively. 
 In the ablation experiments Sec.~\ref{abl: range}, we examine in detail the impact of different values of $w_D$ and $w_I$ on the results.

\noindent \textbf{Cost Volume Construction.} The feature maps $F$ are utilized by using homography warping (i.e., \textit{shift-and-concat (SAC) } \cite{tsai2020attention,chen2021attention,chao2022learning}) to form the coarse cost volume $C_{ref}$. Precisely, the feature maps are shifted along the $u$ or $v$ direction with different predefined disparity samplings and concatenated into the coarse cost volume $C_{coa}$:

 \begin{equation}
\label{eq: coa_cost}
\begin{aligned}
C_{coa} = SAC(F, D_{coa}, I_{coa}),
\end{aligned}
\end{equation}
\noindent  where $I_{coa}$ and $D_{coa}$ are the coarse disparity interval and coarse disparity range. 
Further, we can obtain the coarse disparity map $d_{coa}$ through cost volume aggregation and disparity regression module.
The construction of the refined cost volume $C_{ref}$ differs from that of the coarse cost volume $C_{coa}$ because it does not require shifting features according to the original disparity range. Since the coarse disparity $d_{coa}$ has already been obtained, we only need to refine the disparity further around the range of the coarse disparity map $d_{coa}$, which can reduce the number of disparity samplings and maintain a finer disparity interval. Similarly, we use the \textit{SAC} operation to obtain the refined cost volume $C_{ref}$:

\begin{equation}
\label{eq: ref_cost}
\begin{aligned}
C_{ref} = SAC(F, d_{coa}+D_{ref}, I_{ref}),
\end{aligned}
\end{equation}

\noindent where $d_{coa}$ is the coarse disparity, $I_{ref}$ and $D_{ref}$ are the refined disparity interval and refined disparity range.



\begin{figure}[tb]
  \centering
  \includegraphics[width=\linewidth]{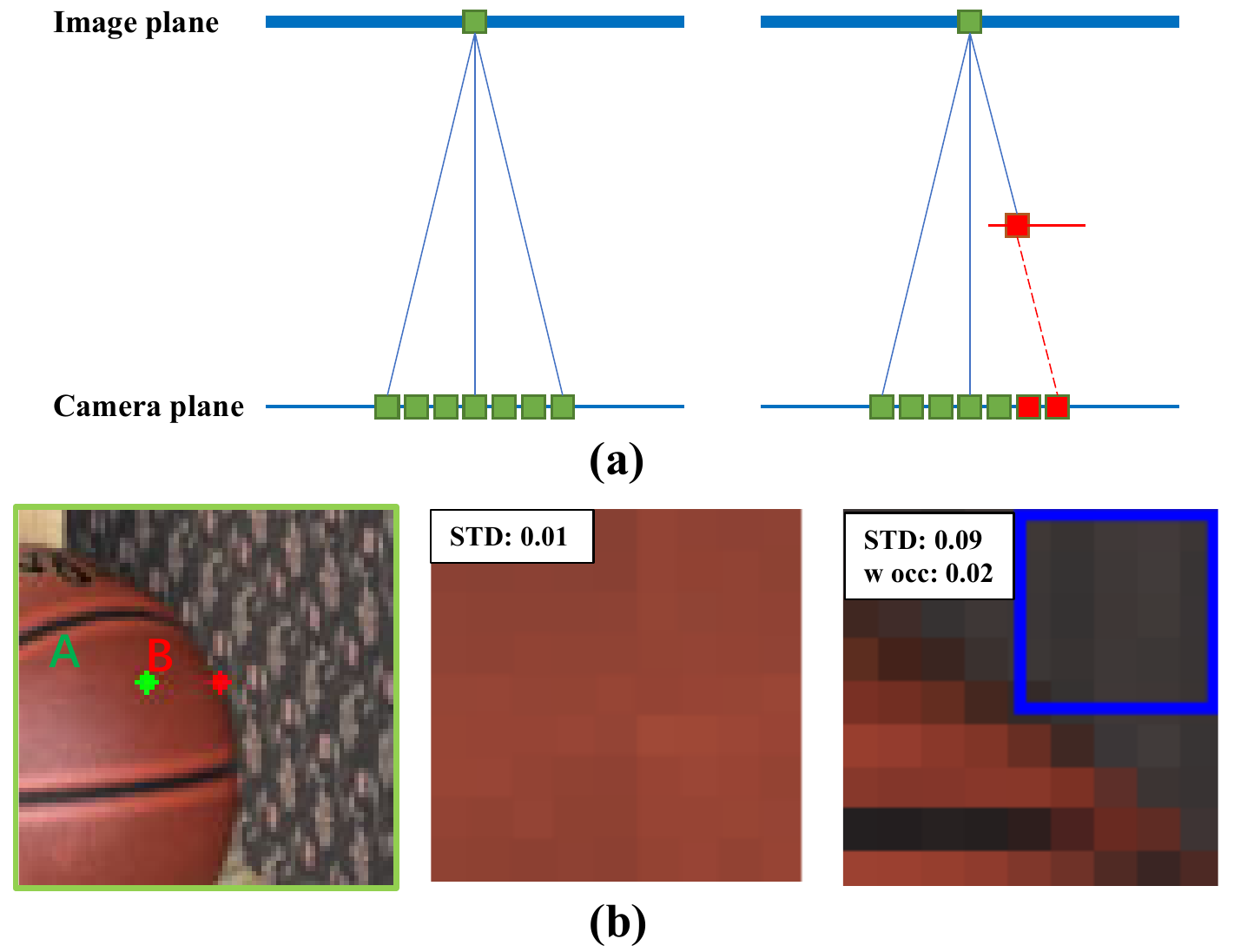}

  \caption{Illustration of the occlusion model. (a) Demonstration of imaging model with occlusion and non-occlusion. The green boxes represent the scene points on the image plane and their projections on the camera plane, while the red boxes indicate the occluders and their projections. (b) Appearances of angular patches of pixels with occlusion and non-occlusion. 
  When refocused on the ground truth depth, point $A$ converges to a point, and the angular patch's standard deviation (STD) is lower. However, the STD value of point $B$ is high due to occlusion. The blue frame highlights the part of the angular patch that follows photo-consistency. Thus, we can utilize an occlusion map to exclude the occlusion pixels, and the angular patch's STD with an occlusion map is also low.}
  \label{fig: occ_model}
\end{figure}

\subsubsection{Occlusion-aware Cost Volume}


For Lambertian scenes, it is commonly assumed that pixels compliance
 with photo-consistency, meaning that when focused to their depth, all viewpoints converge to a single point. However, this assumption does not hold when occlusion occurs, as illustrated  in Fig.~\ref{fig: occ_model}(a). In Fig.~\ref{fig: occ_model}(b), we can see the angular patches of the occluded and unoccluded points when refocused to the ground truth (GT) depth. Point $A$ is unoccluded, so its angular patch has a low standard deviation (STD) value, which demonstrates the photo-consistency constraint. On the other hand, point $B$ has a large STD value even when refocused to GT depth due to occlusion. However, we observed that the pixels in the blue frame still maintain photo-consistency. Therefore, when calculating the STD, we use the occlusion map to exclude the occlusion region; this results in a low STD value.

The above cascade cost volume does not consider the effect of occlusion during the construction process, potentially leading to inaccurate matches. To address this issue, we will introduce an occlusion map to construct an occlusion-aware cost volume.

\begin{figure}[tb]
  \centering
  \includegraphics[width=\linewidth]{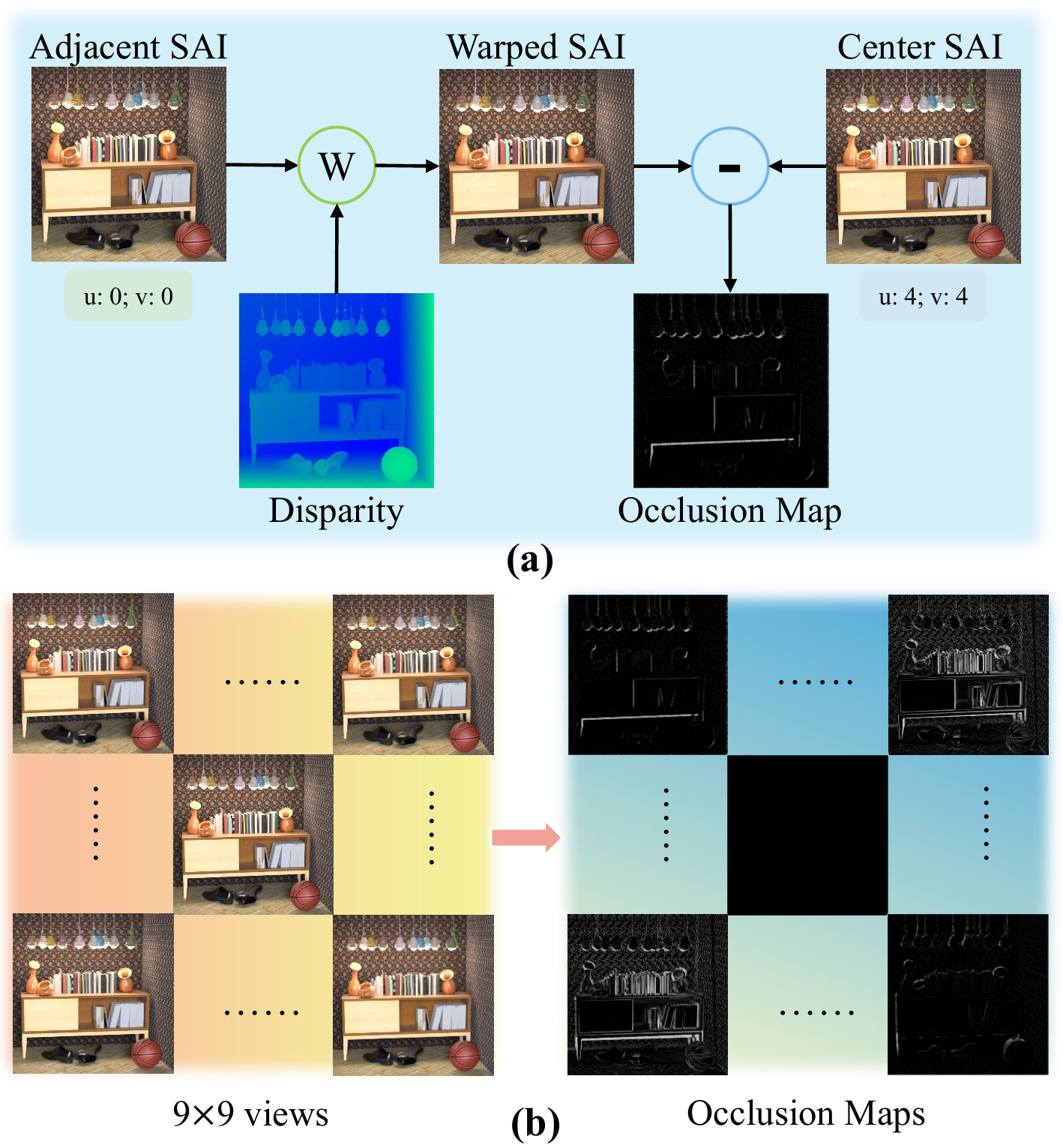}

  \caption{Illustration of the generated occlusion maps on scene \textit{Sideboard}. (a) To calculate the occlusion maps based on the photo-consistency constraints, we use the disparity map to warp the adjacent SAIs to the center view. (b) Visual illustration of occlusion maps of all views. The white regions in the maps represent occlusion.}
  \label{fig: occ}
\end{figure}

\noindent \textbf{Occlusion Maps Generation.} Figure~\ref{fig: occ}(a) illustrates how to generate occlusion maps. First, the surrounding views are warped to the center view:

\begin{equation}
\label{eq: warp}
\begin{aligned}
I_{warp}^{i}=Warp(d_{gt}, I^i), i=1,2,\dots ,U\times V,
\end{aligned}
\end{equation}
where $I_{warp}^{i}$ is the $i$-th warped SAI, $I^i$ is the $i$-th SAI and $d_{gt}$ is the GT disparity map. If there is no occlusion, then the pixels of the warped view are expected to be the same as those of the central view based on the photo-consistency assumption. According to the photo-consistency assumption, the pixels of the warped view should be the same as the central view. Based on the photo-consistency constraint, we can calculate the occlusion maps $M$ for each view: 

\begin{equation}
\label{eq: mask}
\begin{aligned}
M^i=\mathrm{clip} (\left \| I_{center}-I_{warp}^{i} \right \| _1,0,1) 
\end{aligned}
\end{equation}
where $M^i$ is an occlusion map of the $i$-th SAI,  $I_{warp}^{i}$ is the $i$-th warped SAI and $I_{center}$ represents the center SAI. 
Figure \ref{fig: occ} (b) illustrates the occlusion maps for all views, where the occlusion regions are reasonable and consistent with the real case.

\noindent \textbf{Cost Volume with Occlusion Maps.} 
We can represent the importance of each view using $\left \|1-M \right \|_2$ as the dynamic weight. The resulting weighted feature maps $F_w$ can be described as:

\begin{equation}
\label{eq: weighted}
\begin{aligned}
F_{w}^{i} = F^i \odot \left \|1-M^i \right \|_2
\end{aligned}
\end{equation}
where $F_{w}^{i}$ is a weighted feature map of the $i$-th SAI,  $F^i$ is the $i$-th feature map and $M^i$ is a corresponding occlusion map. 
The resulting weighted feature maps $F_w$ take occlusion into account, enabling us to construct an occlusion-aware cost volume using $F_w$. However, obtaining the occlusion maps requires a GT disparity map, which is not always available.
We propose an alternative solution to construct an occlusion-aware cost volume.
Instead of using the GT disparity map, we can use a coarse disparity map and obtain occlusion information when constructing the refined cost volume. 
We can reformulate Eq.~\ref{eq: warp} based on our coarse disparity map $d_{coa}$ to warp the surrounding views:

\begin{equation}
\label{eq: warp2}
\begin{aligned}
I_{warp}^{i}=Warp(d_{coa}, I^i), i=1,2,\dots ,U\times V
\end{aligned}
\end{equation}
where $I_{warp}^{i}$ is the warped image of the $i$-th SAI, $I^i$ is the $i$-th SAI and $d_{coa}$ is the coarse disparity map. We also reformulate Eq.~\ref{eq: ref_cost} to construct the occlusion-aware refined cost volume $C_{ref}$:

\begin{equation}
\label{eq: occ_ref_cost}
\begin{aligned}
C_{ref} = SAC(F_w, d_{coa}+D_{ref}, I_{ref})
\end{aligned}
\end{equation}

\noindent where $d_{coa}$ represents the coarse disparity, $I_{ref}$ and $D_{ref}$ are the refined disparity interval and refined disparity range, respectively. We verified the effectiveness of occlusion maps in the ablation experiment Sec.~\ref{abl: occlusion}.

\subsection{Cascade Disparity Estimation Network}

\noindent \textbf{Feature Extraction.} We follow the methods presented in previous studies \cite{tsai2020attention, chao2022learning}  and adopt the same structure for feature extraction. Each SAI passes through four basic residual blocks \cite{he2016deep} and the spatial pyramid pooling (SPP) module \cite{he2015spatial} to fuse the context information and generate the shared feature map. Specifically, the feature of each SAI is sent to four different  sizes (i.e., 2$\times$2, 4$\times$4, 8$\times$8, and 16$\times$16) of average pooling, and the results are upsampled to the original size. The four levels of features are concatenated with the original feature to generate the output feature map.

\noindent \textbf{Coarse Stage.}
First, the shared feature maps after the SPP module are utilized by the \textit{SAC} operation to construct the coarse cost volume $C_{coar}$. Next, the coarse cost volume $C_{coar}$ is fed into the 3D convolutions and disparity regression module to estimate the normalized probability of each candidate disparity value. Finally, the initial disparity map $d_{coa}$ is calculated by the weighted sum of each disparity $d_k$ with its normalized probability $C_{d_k}$ as the weight:

\begin{equation}
\label{eq: disp}
\begin{aligned}
d_{coa}=\sum_{d_k=D_{coa}^{min}}^{D_{coa}^{max}}d_k \times \mathrm{softmax} (-C_{d_k}),
\end{aligned}
\end{equation}
where $d_{coa}$ represents the coarse disparity map of center SAI, while $D_{coa}^{max}$ and $D_{coa}^{min}$ are the predefined maximum and minimum disparity of the coarse stage, respectively. $d_k$ represents the disparity sampling between $D_{coa}^{max}$ and $D_{coa}^{min}$ based on the disparity interval.

\noindent \textbf{Refined Stage.} First, we warp feature map $F$ based on coarse disparity map $d_{coa}$ to obtain the weighted feature map $F_{w}$,  as described in Eq.~\ref{eq: warp2}. Next, we apply the \textit{SAC} operation to obtain the occlusion-aware refined cost volume $C_{ref}$ at the sub-pixel level, which is beneficial for narrow baselines of LF, as described in Eq.\ref{eq: occ_ref_cost}. Finally, we can generate a refined disparity map $d_{ref}$ similar to Eq.~\ref{eq: disp}:

\begin{equation}
\label{eq: disp2}
\begin{aligned}
d_{ref}=d_{coa} + \sum_{d_k=D_{ref}^{min}}^{D_{ref}^{max}}d_k \times \mathrm{softmax} (-C_{d_k}),
\end{aligned}
\end{equation}
where $d_{coa}$ refers to the coarse disparity map, $d_{ref}$ represents the refined disparity map, $D_{ref}^{max}$ and $D_{ref}^{min}$ are the predefined maximum and minimum disparity of the refined stage, respectively.

\subsection{Loss Functions}

We employ the $L1$ loss as the loss function for each stage, as it is robust to outliers. The total loss $L$ is described as follows: 

\begin{equation}
\label{eq: loss}
\begin{aligned}
L=\lambda_1 \left \| d_{gt}-d_{coa} \right \| _1  + \lambda_2 \left \| d_{gt}-d_{ref} \right \| _1 ,
\end{aligned}
\end{equation}
where $d_{gt}$ represents the GT disparity, $d_{coa}$ and $d_{ref}$ correspond to the coarse stage and refined disparity, and $\lambda_1 = \lambda_2 = 1$, respectively.

\begin{figure*}[!tb]
  \centering
  \includegraphics[width=\linewidth]{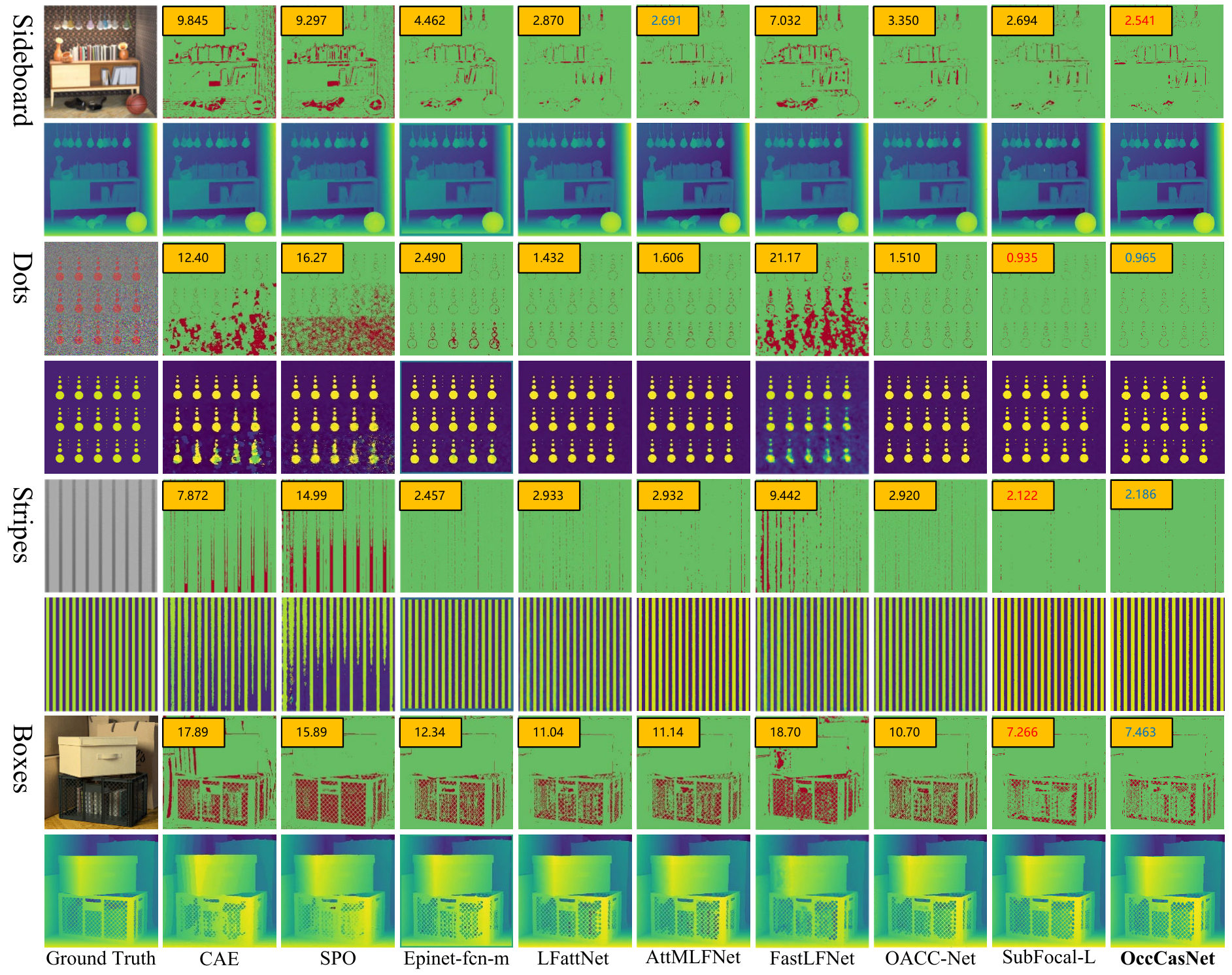}

  \caption{Visual comparisons between our method and state-of-the-art methods on the HCI 4D benchmark\cite{honauer2016dataset} scenes, i.e, \textit{Sideboard}, \textit{Dots}, \textit{Stripes} and \textit{Boxes}, including 
  CAE \cite{park2017robust}, 
  SPO \cite{zhang2016robust}, 
  Epinet-fcn-m \cite{shin2018epinet}, 
  LFattNet \cite{tsai2020attention},
  AttMLFNet \cite{chen2021attention},
  FastLFnet \cite{huang2021fast}, 
  OACC-Net \cite{wang2022occlusion} and SubFocal-L\cite{chao2022learning}, with the corresponding BadPix 0.07 error maps. The best and second-best results are highlighted in red and blue, respectively. Please refer to the supplemental material for additional comparisons.
  }
  \label{fig: comp}
\end{figure*}

\begin{table*}[htb]
\centering
\caption{Quantitative comparison results with state-of-the-art methods on the HCI 4D LF benchmark\cite{honauer2016dataset} regarding BadPix 0.07, BadPix 0.03, BadPix 0.01 and MSE$\times$100. The best and second-best results are highlighted in red and blue, respectively.
}
\renewcommand\arraystretch{1.0}
\resizebox{2.0\columnwidth}{!}{
\begin{tabular}{l|cccccccccccc|c}
\toprule
Method & Backgammon & Dots & Pyramids & Strips & Boxes & Cotton & Dino & Sideboard & Bedroom & Bicycle & Herbs & Origami & Avg. BP 0.07 \\ 
\midrule
CAE\cite{park2017robust} & 3.924 & 12.40 & 1.681 & 7.872 & 17.89 & 3.369 & 4.968 & 9.845 & 5.788 & 11.22 & 9.550 & 10.03 & 8.211 \\
SPO\cite{zhang2016robust} & 3.781 & 16.27 & 0.861 & 14.99 & 15.89 & 2.594 & 2.184 & 9.297 & 4.864 & 10.91 & 8.260 & 11.69 & 8.466 \\
Epinet-fcn-m\cite{shin2018epinet} & 3.501 & 2.490 & \blue{0.159} & 2.457 & 12.34 & 0.447 & 1.207 & 4.462 & 2.299 & 9.614 & 10.96 & 5.807 & 4.646 \\
LFattNet\cite{tsai2020attention} & \blue{3.126} & 1.432 & 0.195 & 2.933 & 11.04 & 0.272 & 0.848 & 2.870 & 2.792 & 9.511 & 5.219 & 4.824 & 3.756 \\
AttMLFNet\cite{chen2021attention} & 3.228 & 1.606 & 0.174 & 2.932 & 11.14 & \red{0.195} & \red{0.440} & \blue{2.691} & \blue{2.074} & 8.837 & 5.426 & 4.406 & 3.596 \\
FastLFnet\cite{huang2021fast} & 5.138 & 21.17 & 0.620 & 9.442 & 18.70 & 0.714 & 2.407 & 7.032 & 4.903 & 15.38 & 10.72 & 12.64 & 9.071 \\
OACC-Net\cite{wang2022occlusion} & 3.931 & 1.510 & \red{0.157} & 2.920 & 10.70 & 0.312 & 0.967 & 3.350 & 2.308 & 8.078 & 6.515 & {4.065} & 3.734 \\
SubFocal-L\cite{chao2022learning} & \red{3.079} & \red{0.935} & 0.253 & \red{2.122} & \red{7.266} & \blue{0.252} & {0.684} & {2.694} & \red{1.882} & \red{6.829} & \blue{3.998} & \red{2.823} & \red{2.735} \\
OccCasNet(Ours) & {3.149} & \blue{0.965} & 0.204 & \blue{2.186} & \blue{7.463} & {0.263} & \blue{0.573} & \red{2.541} & 2.302 & \blue{7.343} & \red{3.896} & \blue{3.400} & \blue{2.859} \\ 
\bottomrule
\toprule
Method & Backgammon & Dots & Pyramids & Strips & Boxes & Cotton & Dino & Sideboard & Bedroom & Bicycle & Herbs & Origami & Avg. BP 0.03 \\ 
\midrule
CAE\cite{park2017robust} & 4.313 & 42.50 & 7.162 & 16.90 & 40.40 & 15.50 & 21.30 & 26.85 & 25.36 & 23.62 & 23.16 & 28.35 & 22.95 \\
SPO\cite{zhang2016robust} & 8.639 & 35.06 & 6.263 & 15.46 & 29.52 & 13.71 & 16.36 & 28.81 & 23.53 & 26.90 & 30.62 & 32.71 & 32.71 \\
Epinet-fcn-m\cite{shin2018epinet} & 5.563 & 9.117 & 0.874 & {2.711} & {18.11} & 2.076 & 3.105 & 10.86 & 6.345 & 16.83 & 25.85 & 13.00 & 9.537 \\
LFattNet\cite{tsai2020attention} & {3.984} & 3.012 & {0.489} & 5.417 & 18.97 & 0.697 & 2.340 & 7.243 & 5.318 & 15.99 & 9.473 & {8.925} & 6.823 \\
AttMLFNet\cite{chen2021attention} & 4.625 & 2.021 & \red{0.429} & 4.743 & 18.65 & \red{0.374} & \red{1.193} & {6.951} & {5.272} & 16.06 & {9.468} & 9.032 & {6.568} \\
FastLFnet\cite{huang2021fast} & 11.41 & 41.11 & 2.193 & 32.60 & 37.45 & 6.785 & 13.27 & 21.62 & 15.92 & 28.45 & 23.39 & 33.65 & 22.32 \\
OACC-Net\cite{wang2022occlusion} & 6.640 & {3.040} & 0.536 & 4.644 & 18.16 & 0.829 & 2.874 & 8.065 & 5.707 & {14.40} & 46.78 & 9.717 & 10.12 \\
SubFocal-L\cite{chao2022learning} & \red{3.651} & \red{1.133} & 0.543 & \red{2.219} & \red{11.41} & \blue{0.501} & {1.735} & \blue{6.246} & \red{3.669} & \red{11.64} & \red{7.238} & \red{6.388} & \red{4.697} \\ 
OccCasNet(Ours) & \blue{3.781} & \blue{1.239} & \blue{0.447} & \blue{2.684} & \blue{14.79} & {0.569} & \blue{1.677} & \red{6.126} & \blue{4.337} & \blue{12.67} & \blue{7.400} & \blue{7.151} & \blue{5.238} \\ 
\bottomrule
\toprule
Method & Backgammon & Dots & Pyramids & Strips & Boxes & Cotton & Dino & Sideboard & Bedroom & Bicycle & Herbs & Origami & Avg. BP 0.01 \\ 
\midrule
CAE\cite{park2017robust} & 17.32 & 83.70 & 27.54 & 39.95 & 72.69 & 59.22 & 61.06 & 56.92 & 68.59 & 59.64 & 59.24 & 64.16 & 55.84 \\
SPO\cite{zhang2016robust} & 49.94 & 58.07 & 79.20 & 21.87 & 73.23 & 69.05 & 69.87 & 73.36 & 72.37 & 71.13 & 86.62 & 75.58 & 66.70 \\
Epinet-fcn-m\cite{shin2018epinet} & 19.43 & 35.61 & 11.42 & {11.77} & 46.09 & 25.72 & 19.39 & 36.49 & 31.82 & 42.83 & 59.93 & 42.21 & 31.90 \\
LFattNet\cite{tsai2020attention} & {11.58} & 15.06 & 2.063 & 18.21 & {37.04} & 3.644 & 12.22 & {20.73} & {13.33} & {31.35} & 19.27 & {22.19} & 17.23 \\
AttMLFNet\cite{chen2021attention} & 13.73 & {10.61} & \blue{1.767} & 15.44 & 37.66 & \red{1.522} & \red{4.559} & 21.56 & 16.18 & 32.71 & {18.84} & 22.45 & {16.42} \\
FastLFnet\cite{huang2021fast} & 39.84 & 68.15 & 22.19 & 63.04 & 71.82 & 49.34 & 56.24 & 61.96 & 52.88 & 59.24 & 59.98 & 72.36 & 56.45\\
OACC-Net\cite{wang2022occlusion} & 21.61 & 21.02 & 3.852 & 15.24 & 43.48 & 10.45 & 22.11 & 28.64 & 21.97 & 32.74 & 86.41 & 32.25 & 28.32 \\
SubFocal-L\cite{chao2022learning} & \blue{7.821} & \red{8.535} & {2.017} & \red{3.992} & \red{29.61} & {3.072} & {9.745} & \blue{18.26} & \red{10.34} & \red{25.66} & \blue{16.65} & \red{18.43} & \red{12.85} \\
OccCasNet(Ours) & \red{7.730} & \blue{9.196} & \red{1.582} & \blue{8.709} & \blue{31.28} & \blue{3.057} & \blue{9.323} & \red{18.08} & \blue{10.62} & \blue{26.58} & \red{16.37} & \blue{19.76}  & \blue{13.52} \\ 
\bottomrule
\toprule
Method & Backgammon & Dots & Pyramids & Strips & Boxes & Cotton & Dino & Sideboard & Bedroom & Bicycle & Herbs & Origami & Avg. MSE \\ 
\midrule
CAE\cite{park2017robust} & 6.074 & 5.082 & 0.048 & 3.556 & 8.424 & 1.506 & 0.382 & 0.876 & 0.234 & 5.135 & 11.67 & 1.778 & 3.730 \\
SPO\cite{zhang2016robust} & 4.587 & 5.238 & 0.043 & 6.955 & 9.107 & 1.313 & 0.310 & 1.024 & 0.209 & 5.570 & 11.23 & 2.032 & 3.968 \\
Epinet-fcn-m\cite{shin2018epinet} & 3.705 & 1.475 & 0.007 & 0.932 & 5.968 & {0.197} & 0.157 & 0.798 & 0.204 & 4.603 & 9.491 & 1.478 & 2.418 \\
LFattNet\cite{tsai2020attention} & \red{3.648} & 1.425 & \blue{0.004} & 0.892 & 3.996 & 0.209 & 0.093 & 0.530 & 0.366 & 3.350 & 6.605 & 1.733 & 1.904 \\
AttMLFNet\cite{chen2021attention} & {3.863} & \red{1.035} & \red{0.003} & \red{0.814} & 3.842 & \red{0.059} & \red{0.045} & \blue{0.398} & \blue{0.129} & 3.082 & {6.374} & 0.991 & 1.720 \\
FastLFnet\cite{huang2021fast} & 3.986 & 3.407 & 0.018 & 0.984 & 4.395 & 0.322 & 0.189 & 0.747 & 0.202 & 4.715 & 8.285 & 2.228 & 2.456 \\
OACC-Net\cite{wang2022occlusion} & 3.938 & 1.418 & \blue{0.004} & \blue{0.845} & {2.892} & \blue{0.162} & {0.083} & 0.542 & 0.148 & {2.907} & 6.561 & \red{0.878} & {1.698} \\
SubFocal-L\cite{chao2022learning} & 3.868 & \blue{1.279} & {0.005} & {0.874} & \red{2.417} & 0.243 & 0.101 & {0.441} & \red{0.125} & \red{2.689} & \blue{6.041} & \blue{0.883} & \blue{1.581} \\
OccCasNet(Ours) & \blue{3.834} & {1.362} & \blue{0.004} & 0.889 & \blue{2.678} & 0.201 & \blue{0.081} & \red{0.397} & {0.135} & \blue{2.705} & \red{5.395} & {0.965} & \red{1.554} \\ \bottomrule
\end{tabular}}
\label{table: quantitative}
\end{table*}

\begin{table*}[tb]
\centering
\caption{Comprehensive comparison with the SubFocal-L\cite{chao2022learning} method on the HCI 4D benchmark\cite{honauer2016dataset}.
The comparison includes several metrics, such as Disparity Number (Disp. Num.), Disparity Interval (Disp. Interv.), Parameters (Params.), FLOPs, Inference Time (Time), BadPix ($\epsilon$), and MSE $\times$100.
FLOPs and Time are calculated at an input LF image size of 9$\times$9$\times$256$\times$256, and the comparison is performed on an NVIDIA RTX 3090 GPU for fairness. 
The BadPix ($\epsilon$) and MSE $\times$100 values represent the average results on the HCI 4D benchmark.
 }
 \renewcommand\arraystretch{1.0}
\resizebox{2.0\columnwidth}{!}{

\begin{tabular}{l|ccccc|cccc}
\toprule
Method & Disp. Num. & Disp. Interv. & Params. & FLOPs & Time & BadPix 0.07 & BadPix 0.03 & BadPix 0.01 & MSE $\times$100 \\ \midrule
SubFocal-L\cite{chao2022learning} & 81 & 1/10 & 5.06M & 53.0T  & 7.13s & 2.735 & 4.697 & 12.85 & 1.581  \\
OccCasNet(Ours) & 33,9 & 1/4,1/8 & 4.79M & 13.2T & 1.13s & 2.859 & 5.238  & 13.52 & 1.554 \\ \bottomrule

\end{tabular}
}
\label{table: detailed}
\end{table*}

\begin{figure*}[tb]
  \centering
  \includegraphics[width=\linewidth]{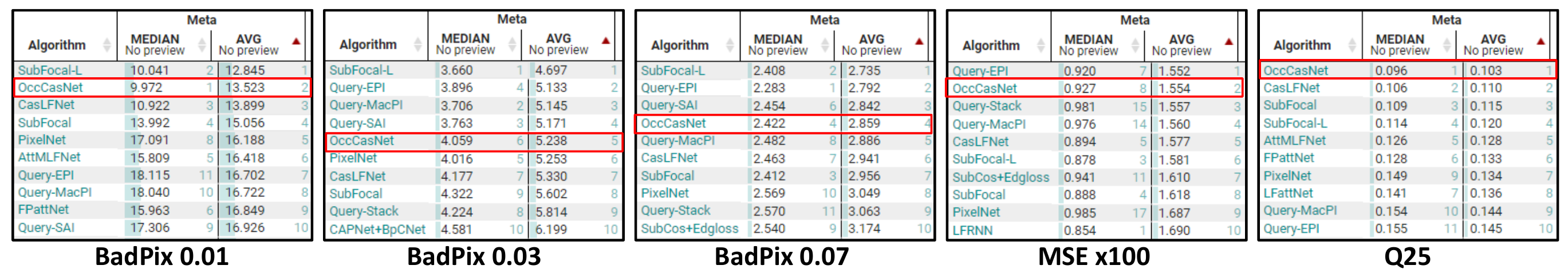}
  
  \caption{The screenshot of HCI 4D LF benchmark\cite{honauer2016dataset} in [https://lightfield-analysis.uni-konstanz.de/] (captured in May 2023). Our method is named “OccCasNet” on the benchmark website.}

  \label{fig: screenshot}
\end{figure*}

\section{Experiments}
\label{sec: experiments}

In this section, we first introduce the datasets and implementation details. Then, we compare our method with state-of-the-art methods. Finally, we conduct extensive ablation experiments to analyze the OccCasNet.

\subsection{Datasets and Implementation Details}

The 4D LF dataset (HCI 4D) \cite{honauer2016dataset} is a widely used synthetic benchmark for evaluating the quality of LF disparity estimation in terms of both quantitative and qualitative performance metrics. The dataset has a spatial resolution of 512$\times$512 and an angular resolution of 9$\times$9. 
Following the setting of previous methods \cite{tsai2020attention, wang2022occlusion, chao2022learning}, we use 16 scenes from the \textit{Additional} subset for training, 8 scenes from the \textit{Training} and \textit{Stratified} subsets for validation, and 4 scenes from the \textit{Test} subset for testing.

We employ a similar network architecture to LFattNet \cite{tsai2020attention}, and adjust the channel of the feature map to reduce the number of parameters. We also adopt the same data augmentation as LFattNet. The hyperparameters $\lambda_1$ and $\lambda_2$ are set to the default value of 1. For the Coarse Stage, we set the disparity range to [-4, 4] with an interval of 1/4, and at the Refined Stage, we set it to [-0.5, 0.5] with an interval of 1/8. We set the batch size to 32, and the grayscale patch size to 32. We use a learning rate of 1e-3 and decay the learning rate by half every 30 epochs. We train the model for 120 epochs using the Adam optimizer based on TensorFlow \cite{abadi2016tensorflow}, and it takes about a week to train on an NVIDIA RTX 3090 GPU.

We evaluate our method using three metrics: Mean square error (MSE $\times$ 100), BadPix($\epsilon $), and Q25. The MSE $\times$100 measures the mean square errors of all pixels, multiplied by 100. BadPix ($\epsilon $) represents the percentage of pixels whose absolute disparity error between the predicted result and the ground truth exceeds a threshold $\epsilon$, commonly set to 0.01, 0.03, and 0.07. Q25 measures the maximum absolute disparity error of the best 25\% pixels, multiplied by 100.

\begin{figure*}[!tb]
  \centering
  \includegraphics[width=\linewidth]{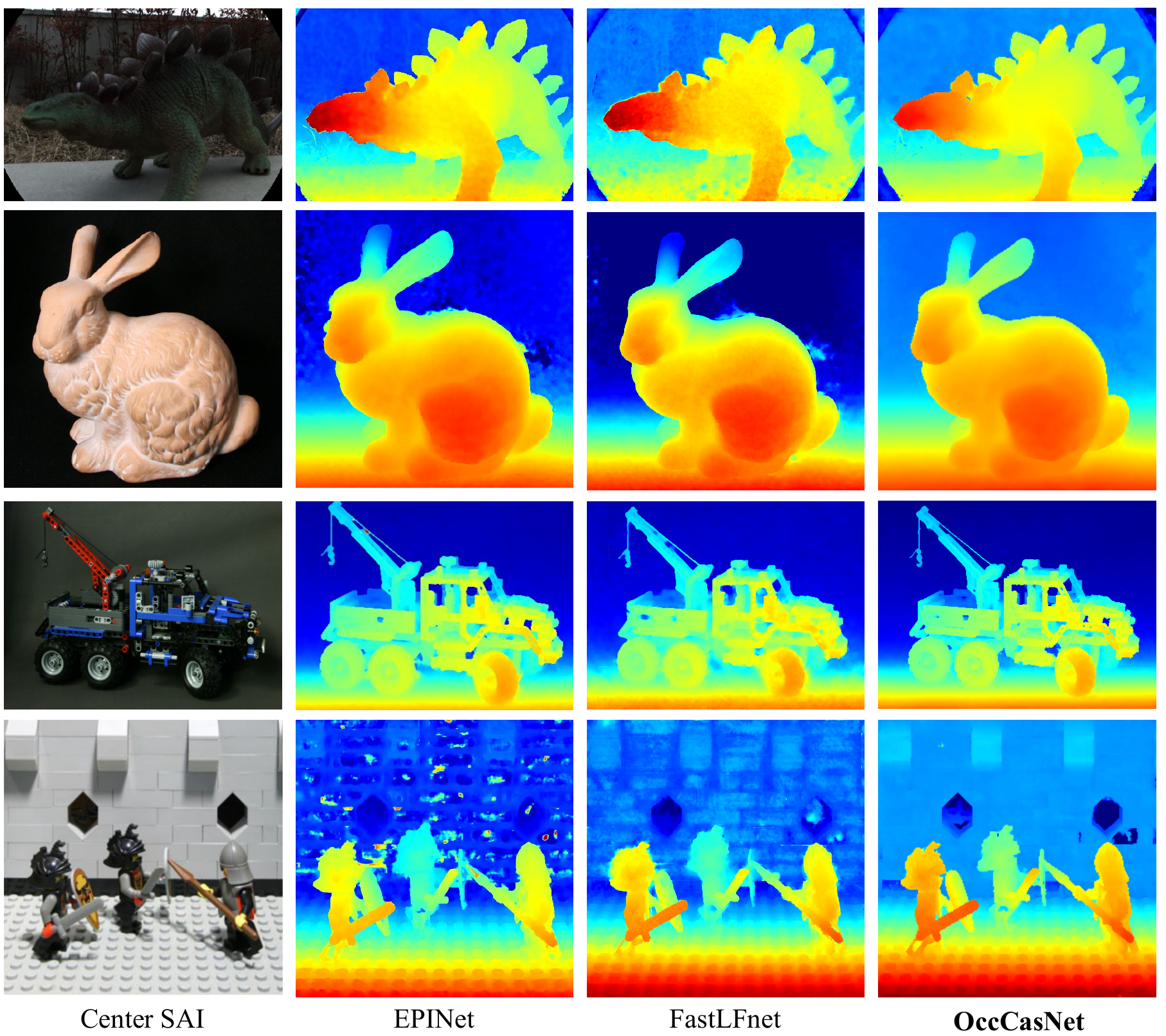}
  \caption{Visual comparisons on real-world scenes, including \textit{Dinosaur}, \textit{Stanford Bunny}, \textit{Truck} and \textit{Knights}. 
  Our results show superior performance compared to EPINet \cite{shin2018epinet} and FastLFnet \cite{huang2021fast}, even in complex scenes like \textit{Truck} and \textit{Knights}. Our method yields more accurate and detailed disparity maps. Please refer to the supplemental material for additional comparisons.
  } 
  \label{fig: real}
\end{figure*}

\subsection{Comparison of State-of-the-art Methods}
\subsubsection{Qualitative Comparison} 
We compare our method with several state-of-art (SOTA) methods, including CAE \cite{park2017robust}, SPO \cite{zhang2016robust}, Epinet-fcn-m\cite{shin2018epinet}, LFattNet\cite{tsai2020attention}, 
AttMLFNet\cite{chen2021attention}, 
FastLFnet \cite{huang2021fast}, OACC-Net\cite{wang2022occlusion} and SubFocal-L\cite{chao2022learning}. 
Figure \ref{fig: comp} shows qualitative comparison results on the scenes of \textit{Sideboard}, \textit{Dots}, \textit{Stripes} and \textit{Boxes}. It is clear that our method and SubFocal-L have less error as compared to other methods, especially in regions with abrupt disparity changes, such as the occlusion regions in the scene \textit{Boxes} and the edge regions in the scene \textit{Dots}.

\subsubsection{Quantitative Comparison} 
We conduct quantitative comparison experiments with 8 SOTA methods\cite{park2017robust, zhang2016robust, shin2018epinet, tsai2020attention, chen2021attention, huang2021fast, wang2022occlusion, chao2022learning}. Table \ref{table: quantitative} shows the comparison results on the HCI 4D benchmark for five metrics: BadPix 0.07, BadPix 0.03, BadPix 0.01, MSE $\times$100, and Q25. Our method is competitive, ranking the top two metrics in most scenes, and ranking first in average MSE $\times$100, and second in average BadPix 0.07, BadPix 0.03, and BadPix 0.01. We have submitted our results to the benchmark website. Figure \ref{fig: screenshot} shows that our method also obtained an overall competitive ranking compared to the methods of published papers as shown in the screenshot of the benchmark website. 

Table \ref{table: detailed} provides a more detailed comparison with SubFocal-L \cite{chao2022learning}, including Disparity Number (Disp. Num.), Disparity Interval (Disp. Interv.), Parameters (Params.), FLOPs, Inference Time (Time),  BadPix ($\epsilon$) and  MSE $\times$100 on the HCI 4D benchmark\cite{honauer2016dataset}. Our method has approximately half the number of disparity samples as compared to SubFocal-L (42 vs. 81), but the sampling interval is essentially the same (1/8 vs. 1/10). The number of parameters is essentially the same for both methods (5.06M vs. 4.79M), while the FLOPs of SubFocal-L are approximately four times higher than ours (53.00T vs. 13.20T), and the inference time is about six times higher (7.13s vs. 1.13s). While our method's BadPix ($\epsilon$) result is slightly higher than SubFocal-L, the MSE $\times$100 is superior to SubFocal-L (1.554 vs. 1.581).

\subsubsection{Performance on Real Scenes}

We compared the performance of our method with EPINet \cite{shin2018epinet} and FastLFnet \cite{huang2021fast} on real scenes captured by a moving camera \cite{vaish2008new} and a Lytro camera \cite{bok2016geometric}. For testing, we used the same model trained on the HCI 4D benchmark, as the GT disparity maps for the real scenes were not available. As shown in Fig.~\ref{fig: real}, EPINet and FastLFnet generated a lot of background noise in many scenes, while our method maintains a clear background. Furthermore, in some complex areas (such as the robotic arm of the scene \textit{Truck}), the disparity maps generated by EPINet and FastLFnet have entangled boundaries, while our method clearly separates them. Our method achieves better overall performance, demonstrating its generalization ability.


\begin{figure}[tb]
  \centering
  \includegraphics[width=\linewidth]{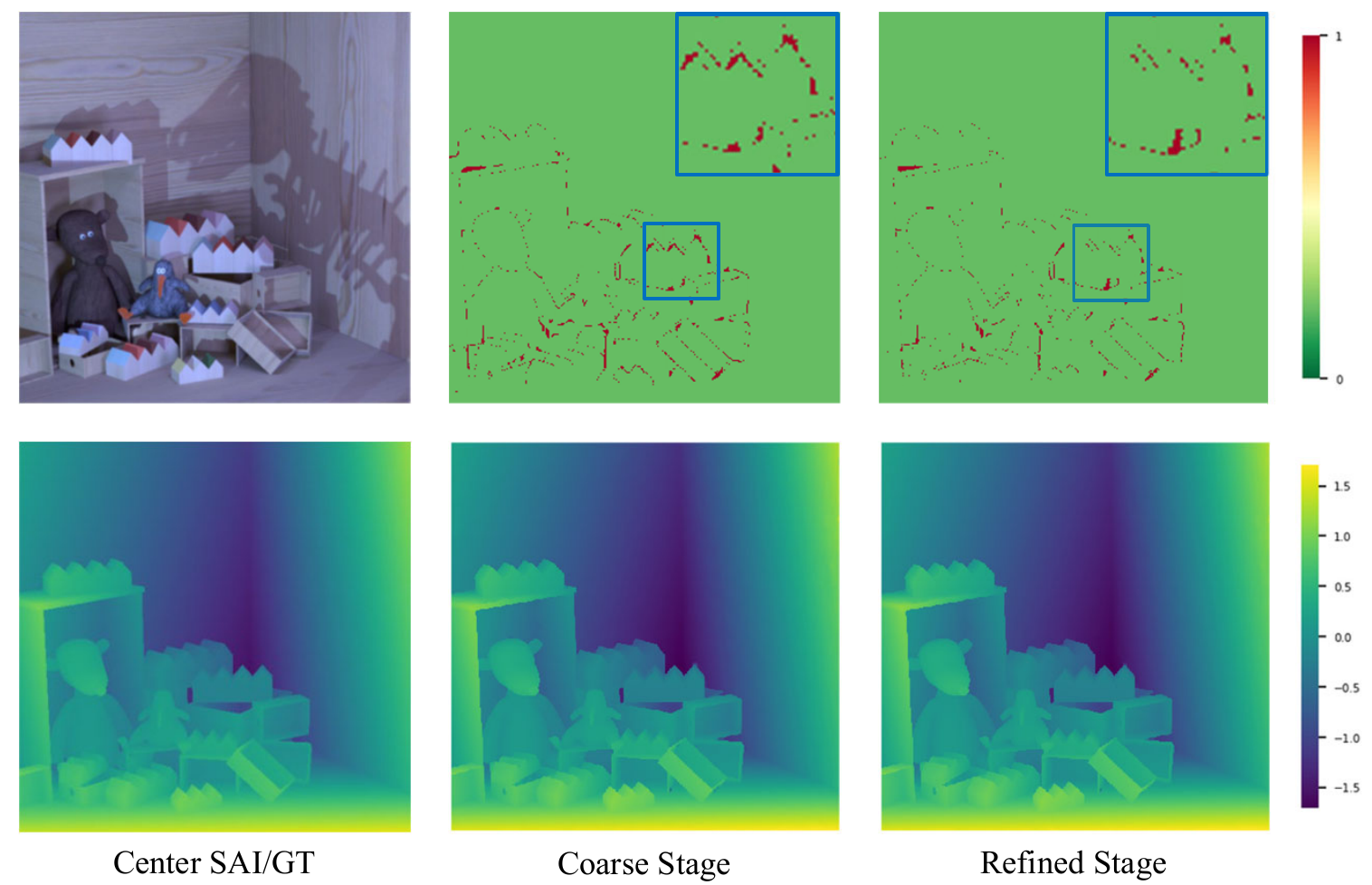}

  \caption{Visual comparison of our method with different stages \textit{Dino} scene. Top-row figures show  the corresponding BadPix 0.07 maps of different stages while the bottom-row figures show the estimated disparity of different stages.}
  \label{fig: abl_stage}
\end{figure}

\begin{table*}[tb]
\centering
\caption{Comprehensive comparison of the cascade cost volume with different settings of disparity number and disparity interval. Comparative metrics include Disparity Number (Disp. Num.), Disparity Interval (Disp. Interv.), Parameters (Params.), FLOPs, BadPix ($\epsilon$), and  MSE $\times$100. FLOPs and Time are calculated at an input LF image size of 9$\times$9$\times$256$\times$256.
The average results for BadPix 0.07 and MSE $\times$100 are obtained from the validation set of the HCI 4D benchmark\cite{honauer2016dataset}.
The default disparity range for stage 1 was -4 to 4.
}
 \renewcommand\arraystretch{1.0}
\resizebox{1.5\columnwidth}{!}{

\begin{tabular}{l|cccc|cc}
\toprule
Method & Disp. Num. & Disp. Interv. & Params. & FLOPs & BadPix 0.07 & MSE $\times$100 \\ \midrule
Cas\_1 & 9 & 1 & 2.41M & 3.11T & 2.804 & 1.164 \\
Cas\_2 & 9,9 & 1,1/8 & 4.79M & 5.87T & \textbf{2.511} & 1.187 \\
Cas\_3 & 9,9,9 & 1,1/8,1/16 & 7.17M & 8.63T & 2.704 & 1.195 \\
Cas\_4 & 9,9,9,9 & 1,1/8,1/16.1/32 & 9.55M & 11.4T & 2.650 & \textbf{1.079} \\ \midrule
Cas\_2 & 9,9 & 1,1/8 & 4.79M & 5.87T & 2.511 & 1.187 \\
Cas\_2-share & 9,9 & 1,1/8 & 2.41M & 5.87T & 5.454 & 1.088 \\
Cas\_2-occ & 9,9 & 1,1/8 & 4.79M & 5.87T & 2.451 & 1.065 \\
Cas\_2-occ-gt  & 9,9 & 1,1/8 & 4.79M & 5.87T & \textbf{2.257} & \textbf{0.792} \\ \midrule
Cas\_2 & 9,9 & 1,1/2 & 4.79M & 5.87T & 2.759 & 1.223 \\
Cas\_2 & 9,9 & 1,1/4 & 4.79M & 5.87T & 2.565 & \textbf{1.077} \\
Cas\_2 & 9,9 & 1,1/8 & 4.79M & 5.87T & \textbf{2.511} & 1.187 \\
Cas\_2 & 9,9 & 1,1/16 & 4.79M & 5.87T & 2.553 & 1.116  \\ \midrule
Cas\_2 & 9,9 & 1,1/8 & 4.79M & 5.87T & 2.651 & 1.076 \\
Cas\_2 & 17,9 & 1/2,1/8 & 4.79M & 8.32T & 2.068 & 0.835 \\
Cas\_2 & 33,9 & 1/4,1/8 & 4.79M & 13.2T & \textbf{1.828} & \textbf{0.780} \\ \midrule
Cas\_2-occ & 33,9 & 1/4,1/8 & 4.79M & 13.2T & \textbf{1.733} & \textbf{0.762} \\
\bottomrule

\end{tabular}
}
\label{table: ablation}
\end{table*}

\begin{figure}[tb]
  \centering
  \includegraphics[width=\linewidth]{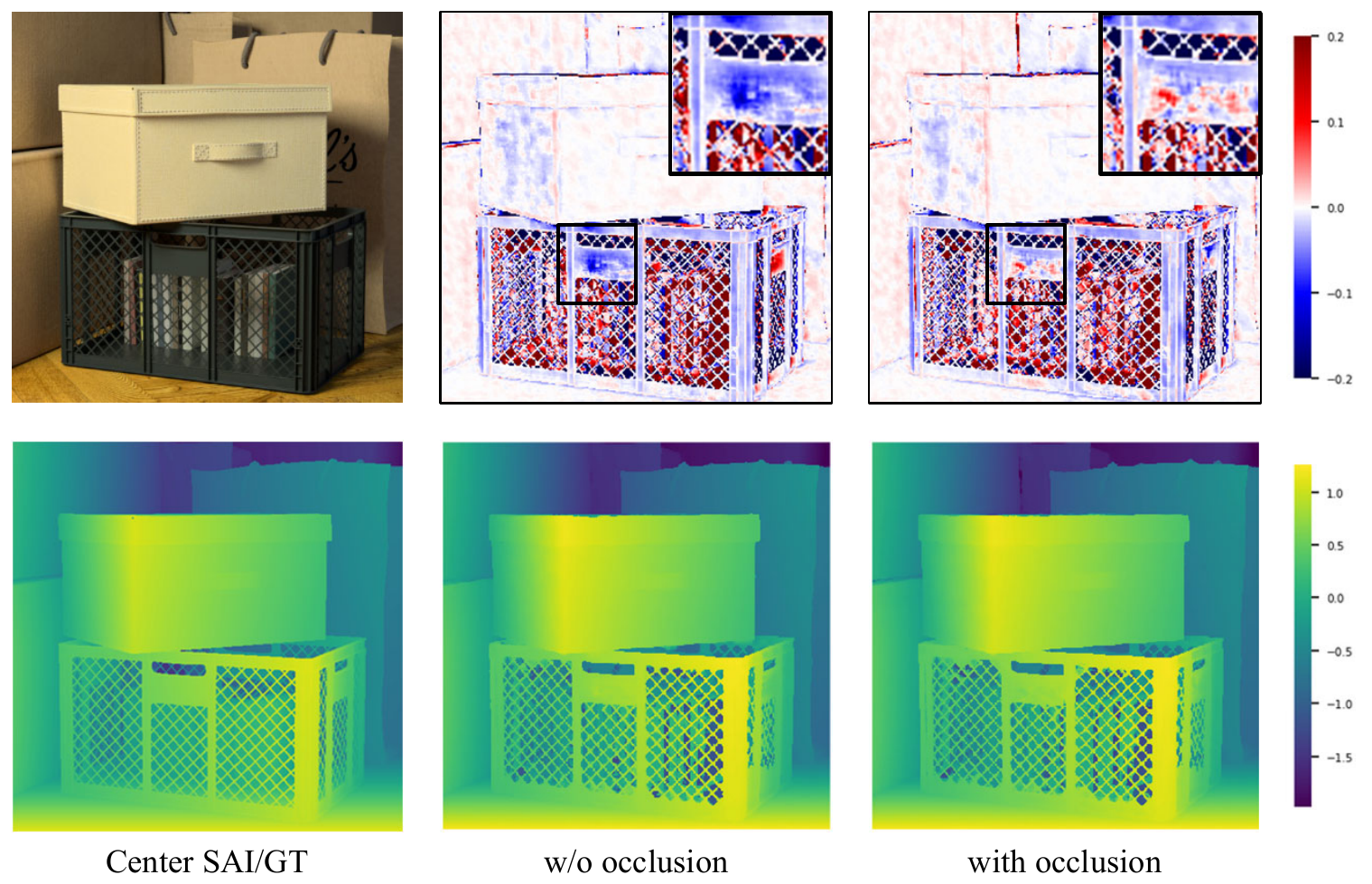}
  \caption{Visual comparison between our method with and without occlusion maps during the construction of the cost volume on the scene \textit{Boxes}. Top-row figures show  the corresponding error maps ( $d_{ref}-d_{gt}$) and the bottom-row figures show the estimated disparity $d_{ref}$.}
  \label{fig: abl_occ}
\end{figure}

\subsection{Ablation Study}

Extensive ablation studies are performed to validate the improved accuracy and efficiency of our approach. Specifically, we study the effects of cascade stage number, disparity range and disparity interval, parameter sharing in cost volume regularization, and occlusion maps in cost volume construction.

\subsubsection{Cascade Stage Number} The quantitative results with different stage numbers are shown in Table\ref{table: ablation}. Cas\_$i$ stands for using the $i$-th stage to predict disparity. We find that as the number of stages increases, the BadPix 0.07 and MSE $\times$100 metrics first decrease significantly and then stabilize. Specifically, by comparing the results of stage 1 and stage 2, the BadPix 0.07 decreases from 2.804 to 2.511.
However, as the number of stages increases, the parameters and FLOPs of the model also increase, as seen in the parameter comparison (4.79M vs. 9.55M) and FLOPs comparison (5.87T vs. 11.4T) between stage 2 and stage 4. Considering the trade-off between speed and accuracy, we ultimately choose a two-stage cascade network. From Fig.~\ref{fig: abl_stage}, it can be observed that stage 1 (coarse stage) can obtain a better coarse disparity, while stage 2 (refined stage) further improves the accuracy of the disparity map. 

\subsubsection{Disparity Range and Disparity Interval}
\label{abl: range}
The choice of disparity range and disparity interval has a significant influence on the accuracy and speed of the model. The disparity range for stage 1 (coarse stage) can be determined based on the actual scene, while the disparity interval should be selected according to the desired trade-off between accuracy and speed.
The optimal disparity range and interval for stage 2 (refined stage) are also determined through experiments.

We adopt a control variable approach to determine the appropriate disparity range and interval for the two-stage cascade network. We first fix the disparity range of stage 1 to be -4 to 4, with a disparity interval of 1 and a disparity number of 9. 
Then, we adjust the disparity range and disparity interval of stage 2 and evaluate the performance to find the optimal values. 
As shown in Table \ref{table: ablation}, we find that the optimal BadPix 0.07 metric is achieved when the number of disparities in stage 2 is 9 and the disparity interval is 1/8. We also fix the disparity range and disparity interval of stage 2 and vary the disparity interval of stage 1. We can see from Table \ref{table: ablation} that as the stage 1 disparity interval decreases, the overall performance significantly improves, while the FLOPs increase correspondingly. The selection of stage 1 disparity interval can be determined based on actual needs. In this paper, we choose the stage 1 disparity interval as 33 for higher accuracy.

\subsubsection{Parameter Sharing in Cost Volume Regularization}

In addition, we conducted a study to determine whether the cost volume in different stages could be parameter-shared. As shown in Table \ref{table: ablation}, compared to the shared parameter model, the number of parameters of the model is reduced by half (4.79M vs. 2.41M). However, the BadPix 0.07 metric increases significantly (2.511 vs. 5.454). This experiment demonstrates that the parameters of different stages need to be learned separately, possibly due to the different settings of disparity number and disparity intervals in different stages.

\subsubsection{Occlusion Maps in Cost Volume Construction}
\label{abl: occlusion}
When constructing the occlusion-aware refined cost volume, we introduce the occlusion maps calculated by the coarse disparity map, which is predicted by the coarse stage. We conducted experiments to verify the effectiveness of occlusion maps. Table \ref{table: ablation} shows the quantitative results with and without the occlusion maps in the refined stage. Our method achieves less error when using the occlusion maps to guide the construction of the refined cost volume, compared to when the occlusion maps are not used. We also conducted an upper-bound experiment using the real disparity map to calculate the occlusion maps, and the results show that the error is further reduced. It can be seen from Fig.~\ref{fig: abl_occ} that when constructing cost volume with occlusion maps compared without occlusion maps, the error at the occluded edge will be less than without occlusion maps, confirming that our occlusion-aware cost volume can alleviate the occlusion problem in LF disparity estimation.


\begin{figure}[tb]
  \centering
  \includegraphics[width=\linewidth]{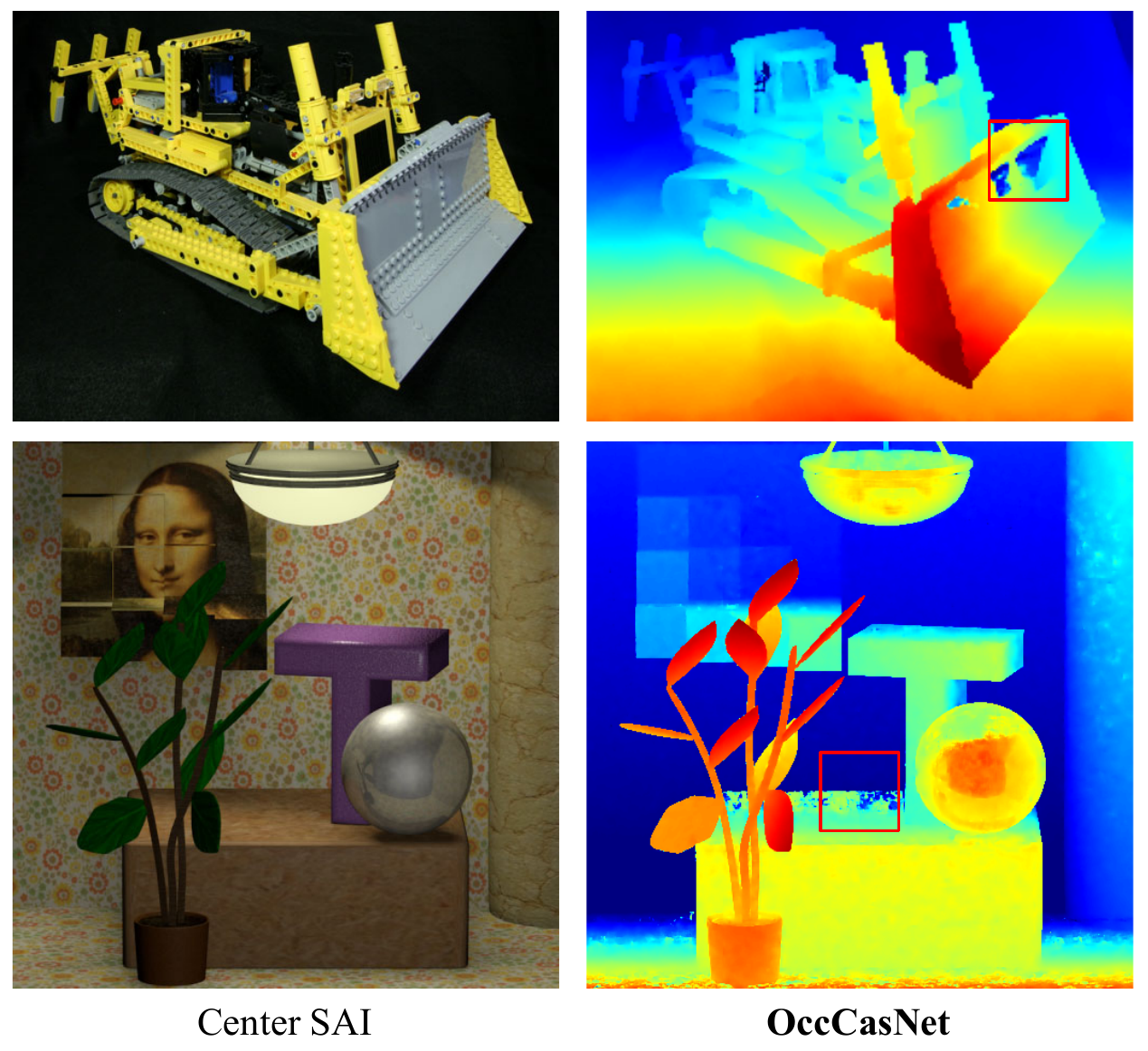}

  \caption{Failure cases on challenge scenes, i.e, \textit{Lego Bulldozer} and \textit{monasRoom}.}
  \label{fig: fail}
\end{figure}

\section{Conclusion and Discussion}
\label{sec:conclusion}

In this paper, we propose a novel and effective method, i.e., OccCasNet, for LF disparity estimation. We construct a cascade cost volume at a finer level in a coarse-to-fine manner. On the other hand, the occlusion maps are introduced to guide the construction of occlusion-aware refined cost volume. Extensive experiments demonstrate the effectiveness of our method. Compared with state-of-art methods, our method can increase the speed while maintaining high accuracy.

Despite the progress shown by our method as compared to state-of-the-art methods, there are still some limitations. First, our method only considers occlusion and does not address other complex situations, such as weak textures and highlights, as shown in Fig.~\ref{fig: fail}. To overcome this limitation, we plan to expand the receptive field through the design scheme to alleviate the situation of weak textures. For the specular area, we can first perform specular separation and then perform disparity estimation.
Second, the speed of our method is still not fast enough, and the number of parameters is too large. We attribute this to the construction of the sub-pixel cost volume and the corresponding cost volume aggregation. To address this limitation, we plan to design a cost volume construction scheme based on convolution or parallel shift, inspired by OACC-Net\cite{wang2022occlusion}. We also aim to explore the use of cost volume metrics, such as mean and variance, to further reduce the amount of calculation and running time.
We hope that our method can inspire further research in LF disparity estimation and contribute to the development of more efficient and accurate methods.



\bibliographystyle{IEEEtran}
\bibliography{reference}

\clearpage

\begin{appendix} 

\section{Appendix}

Our OccCasNet is described in detail in Sec. \ref{sup:network}. The 4D light field (LF) benchmark is further compared in Sec. \ref{sub:results}. Section \ref{sub:visual} displays additional visual results obtained using various techniques on other LF datasets \cite{le2018light,rerabek2016new,vaish2008new,wanner2013datasets}.

\subsection{Details of our OccCaNet}
\label{sup:network}

\begin{figure}[htb]
  \centering
  \includegraphics[width=\linewidth]{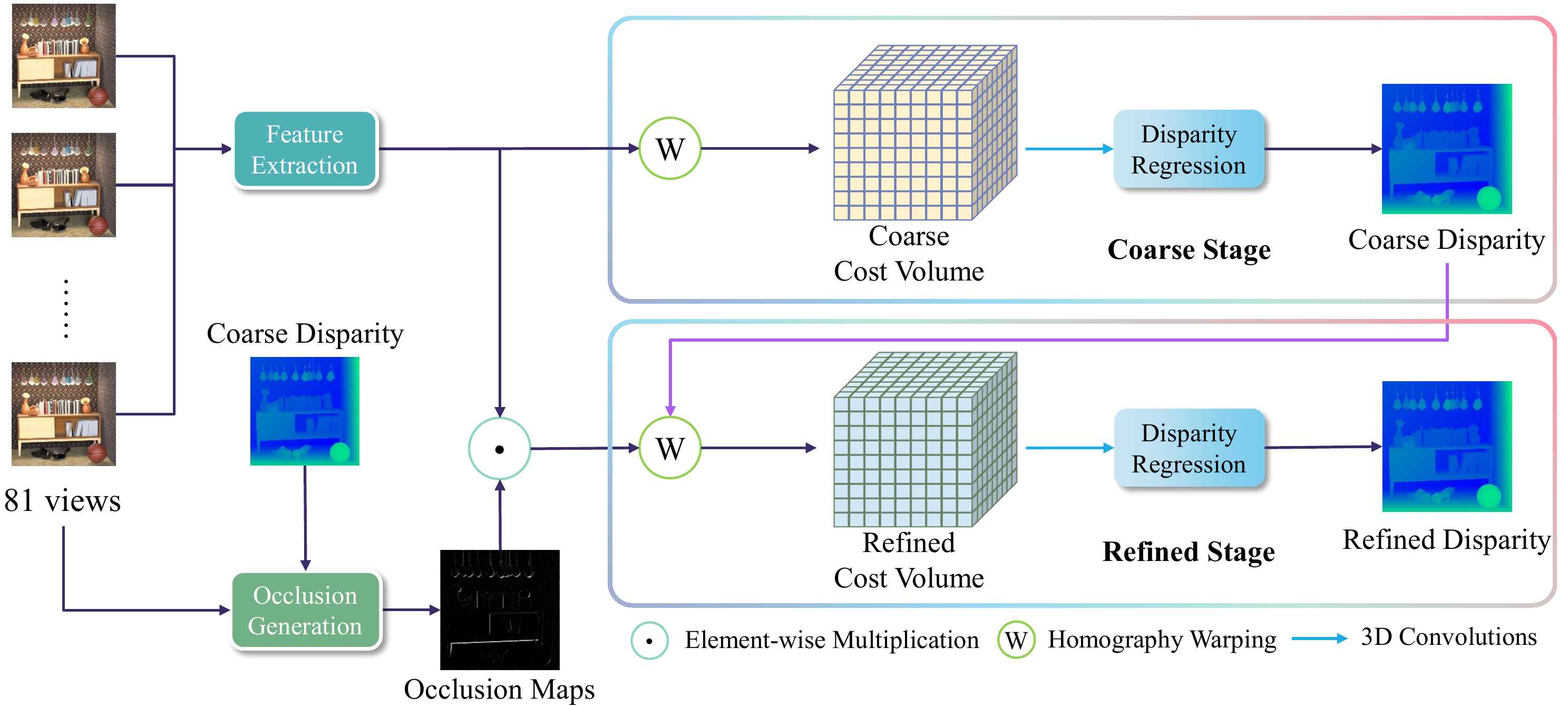}

  \caption{The feature extraction module is utilized to extract the features of each SAI and form the shared feature map. The coarse disparity estimation network is adopted to generate the coarse disparity map. Additionally, the occlusion generation module is used to calculate the occlusion maps. Finally, the refined disparity estimation network takes both the coarse disparity and occlusion maps as inputs to generate a refined disparity map.}
  \label{fig: network2}
\end{figure}

\begin{table}[tb]
\centering
\caption{The detailed architecture of our OccCaNet. 
$Conv2D$, $Conv3D$, $ResBlock3D$ represents 2D convolution, 3D convolution and 3D residual block, respectively. $H$ and $W$ are the height and width. $M$ denotes the number of views (e.g.,$M=9\times9=81$), and $D$ denotes the number of disparity candidates (e.g.,$D=9$).}
\label{table: network}
\renewcommand\arraystretch{1.5}
\resizebox{\columnwidth}{!}{
\begin{tabular}{|cccc|}
\hline
\multicolumn{1}{|c|}{Layers} & \multicolumn{1}{c|}{Kernel Size} & \multicolumn{1}{c|}{Input Size} & Output Size \\ \hline
\multicolumn{4}{|c|}{Feature Extraction} \\ \hline
\multicolumn{1}{|c|}{Conv2D\_1} & \multicolumn{1}{c|}{3$\times$3} & \multicolumn{1}{c|}{M$\times$(H$\times$W$\times$1)} & M$\times$(H$\times$W$\times$4) \\ \hline
\multicolumn{1}{|c|}{Conv2D\_2} & \multicolumn{1}{c|}{3$\times$3} & \multicolumn{1}{c|}{M$\times$(H$\times$W$\times$4)} & M$\times$(H$\times$W$\times$4) \\ \hline
\multicolumn{1}{|c|}{SPP Module} & \multicolumn{1}{c|}{\begin{tabular}[c]{@{}c@{}}2$\times$2\\ 4$\times$4\\ 8$\times$8\\ 16$\times$16\end{tabular}} & \multicolumn{1}{c|}{M$\times$(H$\times$W$\times$4)} & M$\times$(H$\times$W$\times$4) \\ \hline
\multicolumn{4}{|c|}{Coarse Cost Volume Construction} \\ \hline
\multicolumn{1}{|c|}{SAC} & \multicolumn{1}{c|}{-} & \multicolumn{1}{c|}{M$\times$(H$\times$W$\times$4)} & D$\times$H$\times$W$\times$(4$\times$M) \\ \hline
\multicolumn{1}{|c|}{Channel Attention} & \multicolumn{1}{c|}{1x1x1} & \multicolumn{1}{c|}{D$\times$H$\times$W$\times$(4$\times$M)} & D$\times$H$\times$W$\times$(4$\times$M) \\ \hline
\multicolumn{4}{|c|}{Coarse Cost Aggregation} \\ \hline
\multicolumn{1}{|c|}{Conv3D\_1} & \multicolumn{1}{c|}{3$\times$3$\times$3} & \multicolumn{1}{c|}{D$\times$H$\times$W$\times$(4$\times$M)} & DxH$\times$W$\times$96 \\ \hline
\multicolumn{1}{|c|}{Conv3D\_2} & \multicolumn{1}{c|}{3$\times$3$\times$3} & \multicolumn{1}{c|}{DxH$\times$W$\times$96} & DxH$\times$W$\times$96 \\ \hline
\multicolumn{1}{|c|}{ResBlock3D $\times$2} & \multicolumn{1}{c|}{\begin{tabular}[c]{@{}c@{}}3$\times$3$\times$3\\ 3$\times$3x3\end{tabular}} & \multicolumn{1}{c|}{DxH$\times$W$\times$96} & DxH$\times$W$\times$96 \\ \hline
\multicolumn{1}{|c|}{Conv3D\_3} & \multicolumn{1}{c|}{3$\times$3$\times$3} & \multicolumn{1}{c|}{D$\times$H$\times$W$\times$96} & DxH$\times$W$\times$96 \\ \hline
\multicolumn{1}{|c|}{Cost} & \multicolumn{1}{c|}{3$\times$3$\times$3} & \multicolumn{1}{c|}{D$\times$H$\times$W$\times$96} & DxH$\times$W$\times$1 \\ \hline
\multicolumn{1}{|c|}{Squeeze\&Transpose} & \multicolumn{1}{c|}{-} & \multicolumn{1}{c|}{D$\times$H$\times$W$\times$1} & H$\times$W$\times$D \\ \hline
\multicolumn{4}{|c|}{Coarse Disparity Regression} \\ \hline
\multicolumn{1}{|c|}{Softmax} & \multicolumn{1}{c|}{-} & \multicolumn{1}{c|}{H$\times$W$\times$D} & H$\times$W$\times$D \\ \hline
\multicolumn{1}{|c|}{Regress} & \multicolumn{1}{c|}{-} & \multicolumn{1}{c|}{H$\times$W$\times$D} & H$\times$W$\times$1 \\ \hline
\multicolumn{4}{|c|}{Occlusion Generation} \\ \hline
\multicolumn{1}{|c|}{Warp} & \multicolumn{1}{c|}{-} & \multicolumn{1}{c|}{H$\times$W$\times$D, H$\times$W$\times$1} & H$\times$W$\times$D \\ \hline

\multicolumn{4}{|c|}{Refined Cost Volume Construction} \\ \hline
\multicolumn{1}{|c|}{Warp\&SAC} & \multicolumn{1}{c|}{-} & \multicolumn{1}{c|}{M$\times$(H$\times$W$\times$4), H$\times$W$\times$D} & D$\times$H$\times$W$\times$(4$\times$M) \\ \hline
\multicolumn{1}{|c|}{Channel Attention} & \multicolumn{1}{c|}{1x1x1} & \multicolumn{1}{c|}{D$\times$H$\times$W$\times$(4$\times$M)} & D$\times$H$\times$W$\times$(4$\times$M) \\ \hline
\multicolumn{4}{|c|}{Refined Cost Aggregation} \\ \hline
\multicolumn{1}{|c|}{Conv3D\_1} & \multicolumn{1}{c|}{3$\times$3$\times$3} & \multicolumn{1}{c|}{D$\times$H$\times$W$\times$(4$\times$M)} & DxH$\times$W$\times$96 \\ \hline
\multicolumn{1}{|c|}{Conv3D\_2} & \multicolumn{1}{c|}{3$\times$3$\times$3} & \multicolumn{1}{c|}{DxH$\times$W$\times$96} & DxH$\times$W$\times$96 \\ \hline
\multicolumn{1}{|c|}{ResBlock3D $\times$2} & \multicolumn{1}{c|}{\begin{tabular}[c]{@{}c@{}}3$\times$3$\times$3\\ 3$\times$3x3\end{tabular}} & \multicolumn{1}{c|}{DxH$\times$W$\times$96} & DxH$\times$W$\times$96 \\ \hline
\multicolumn{1}{|c|}{Conv3D\_3} & \multicolumn{1}{c|}{3$\times$3$\times$3} & \multicolumn{1}{c|}{D$\times$H$\times$W$\times$96} & DxH$\times$W$\times$96 \\ \hline
\multicolumn{1}{|c|}{Cost} & \multicolumn{1}{c|}{3$\times$3$\times$3} & \multicolumn{1}{c|}{D$\times$H$\times$W$\times$96} & DxH$\times$W$\times$1 \\ \hline
\multicolumn{1}{|c|}{Squeeze\&Transpose} & \multicolumn{1}{c|}{-} & \multicolumn{1}{c|}{D$\times$H$\times$W$\times$1} & H$\times$W$\times$D \\ \hline
\multicolumn{4}{|c|}{Refined Disparity Regression} \\ \hline
\multicolumn{1}{|c|}{Softmax} & \multicolumn{1}{c|}{-} & \multicolumn{1}{c|}{H$\times$W$\times$D} & H$\times$W$\times$D \\ \hline
\multicolumn{1}{|c|}{Regress} & \multicolumn{1}{c|}{-} & \multicolumn{1}{c|}{H$\times$W$\times$D} & H$\times$W$\times$1 \\ \hline
\end{tabular}
}

\end{table}

Fig. \ref{fig: network2} shows the pipeline of our OccCaNet, including feature extraction, coarse cost volume construction, coarse cost aggregation, coarse disparity regression, occlusion generation, refined cost volume construction, refined cost aggregation, and refined disparity regression. The detailed structure of our OccCaNet is shown in Table \ref{table: network}. We describe the details of each module in detail below.

\subsubsection{Feature Extraction} 
First, two 3 $\times$ 3 convolutions (i.e., \textit{Conv2D\_1} and \textit{Conv2D\_2)} are used to extract the initial feature with a channel of 4. Then, we use the SPP module to extract multi-scale features. SPP module is set as follows:
\begin{enumerate}

\item  Four average pooling operations at different scales are used to compress the features. The sizes of the average pooling blocks are 2 $\times$ 2, 4 $\times$ 4, 8 $\times$ 8, and 16 $\times$ 16. 
\item  A 1 $\times$ 1 convolution layer is used for reducing the feature dimension for each scale. 
\item  Bilinear interpolation is adopted to upsample these low-dimensional feature maps to the same size.
\item  Concatenating the feature maps of all levels as the output feature map of the SPP module.

\end{enumerate}

\subsubsection{Coarse/Refined Cost Volume Construction}

The feature maps $F$ are utilized by using homography warping (i.e., \textit{shift-and-concat (SAC) } \cite{tsai2020attention,chen2021attention,chao2022learning}) to form the coarse cost volume $C_{ref}$. Precisely, the feature maps are shifted along the $u$ or $v$ direction with different predefined disparity samplings and concatenated into the coarse cost volume $C_{coa}$.

Further, we can obtain the coarse disparity map $d_{coa}$ through cost volume aggregation and disparity regression module.
The construction of the refined cost volume $C_{ref}$ differs from that of the coarse cost volume $C_{coa}$ because it does not require shifting features according to the original disparity range. Since the coarse disparity $d_{coa}$ has already been obtained, we only need to refine the disparity further around the range of the coarse disparity map $d_{coa}$, which can reduce the number of disparity samplings and maintain a finer disparity interval. Similarly, we use the \textit{SAC} operation to obtain the refined cost volume $C_{ref}$

\subsubsection{Coarse/Refined Cost Aggregation}

Our architecture consists of eight $3\times3\times3$ convolutional layers, with two residual blocks from the third to the sixth 3D convolutional layers. Then We use the squeeze and transpose operation to adjust dimensions. Finally,  the output cost volume of cost aggregation is 3D tensor $H\times W\times D$. 

\subsubsection{Coarse/Refined Disparity Regression}
We use the softmax operation to calculate the disparity distribution $H\times W\times D$. Then, the final disparity map $H\times W\times 1$ is calculated by the weighted sum of each disparity distribution  with its normalized probability as the weight.

\subsection{Results on the 4D LF Benchmark}
\label{sub:results}
\begin{figure*}[htb]
  \centering
  \includegraphics[width=0.95 \linewidth]{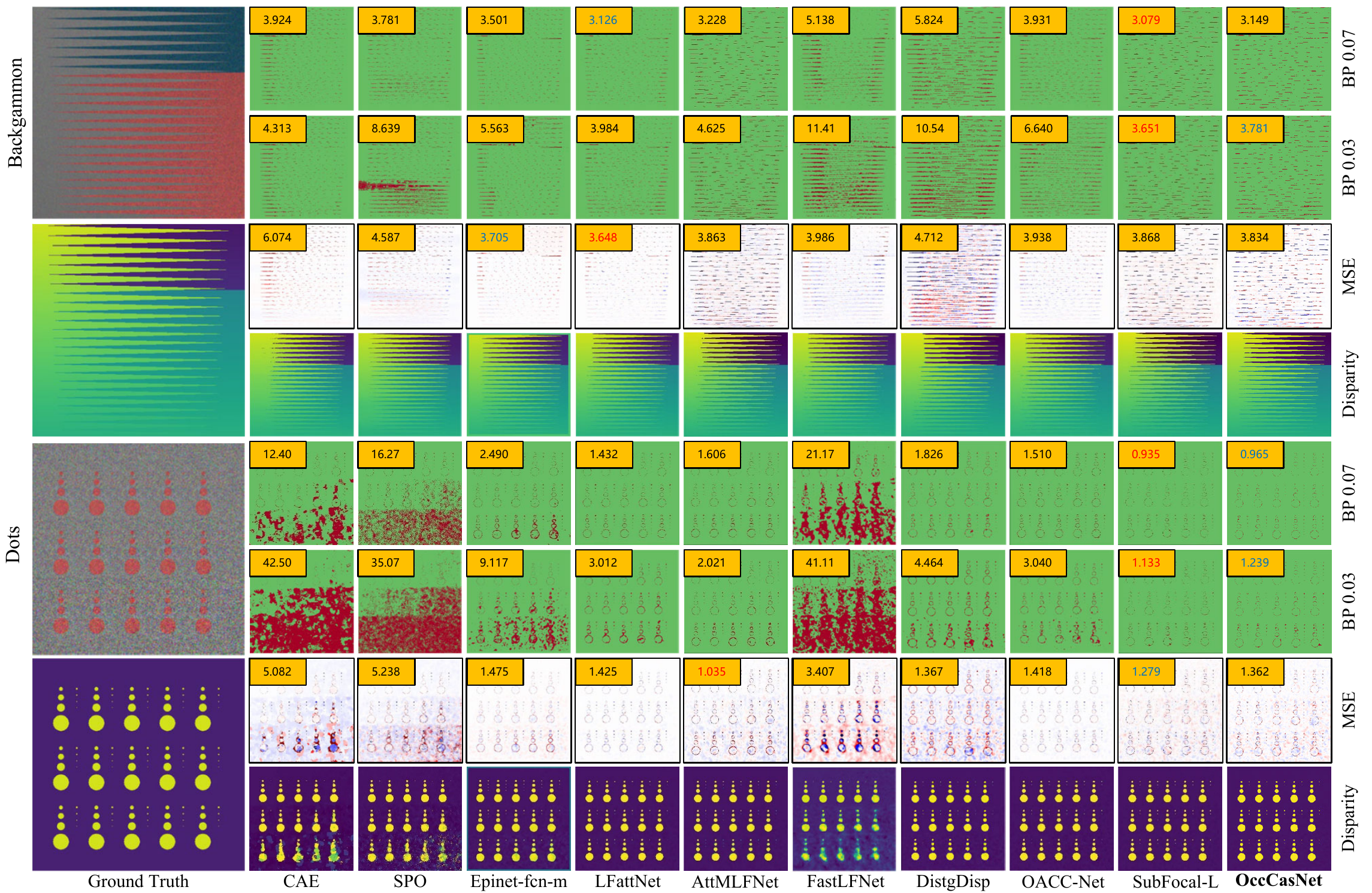}
  \includegraphics[width=0.95 \linewidth]{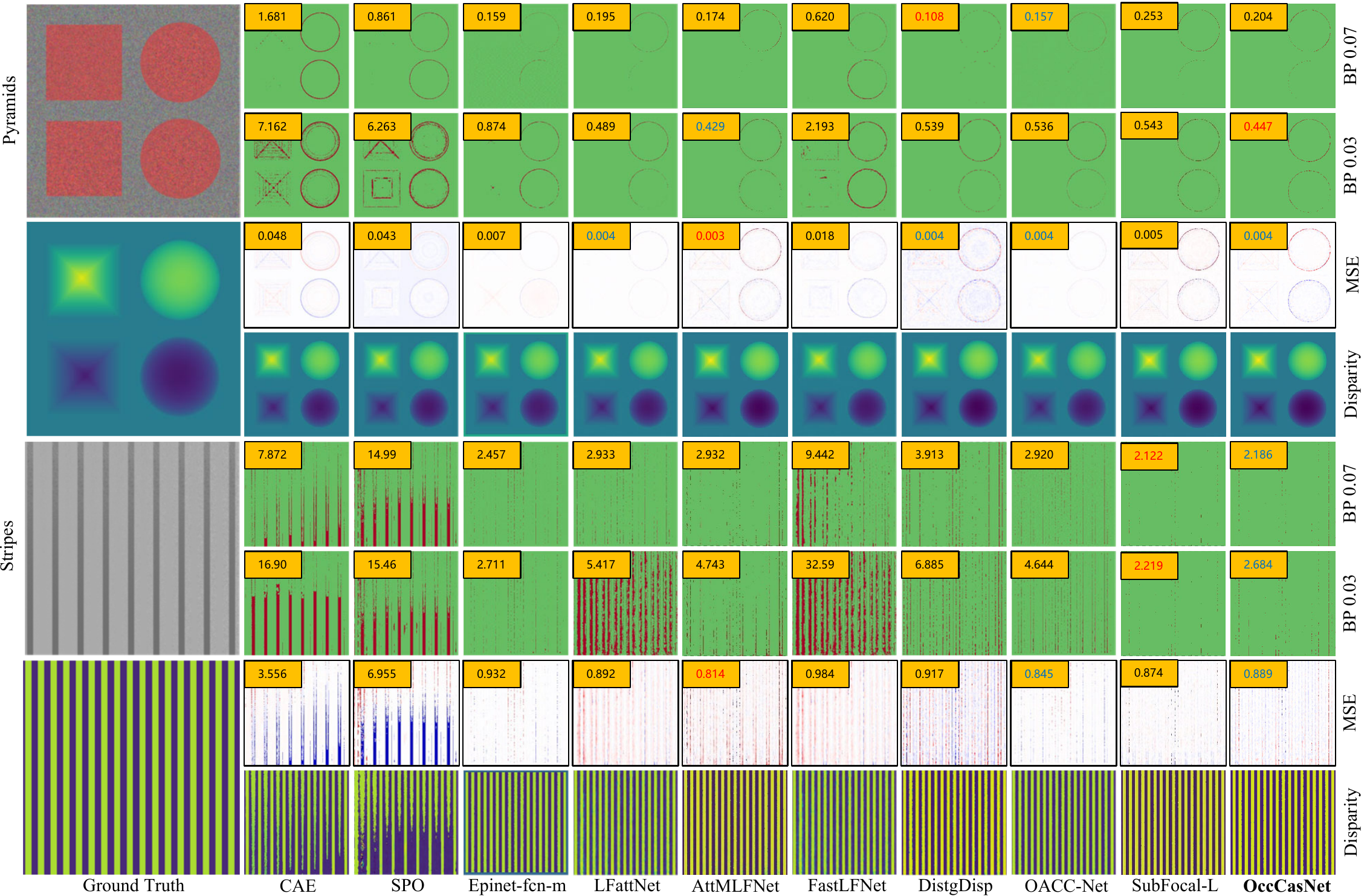}
  \caption{Visual comparisons of disparity and error maps on validation scenes \textit{backgammon}, \textit{dots}, \textit{pyramids}, and \textit{stripes} \cite{rerabek2016new}. Corresponding quantitative scores (BadPix0.07, BadPix0.03, and MSE) are reported on the top-left corner of each error map.
  }

  \label{fig: hci_1}
\end{figure*}

\begin{figure*}[htb]
  \centering
  \includegraphics[width=0.95\linewidth]{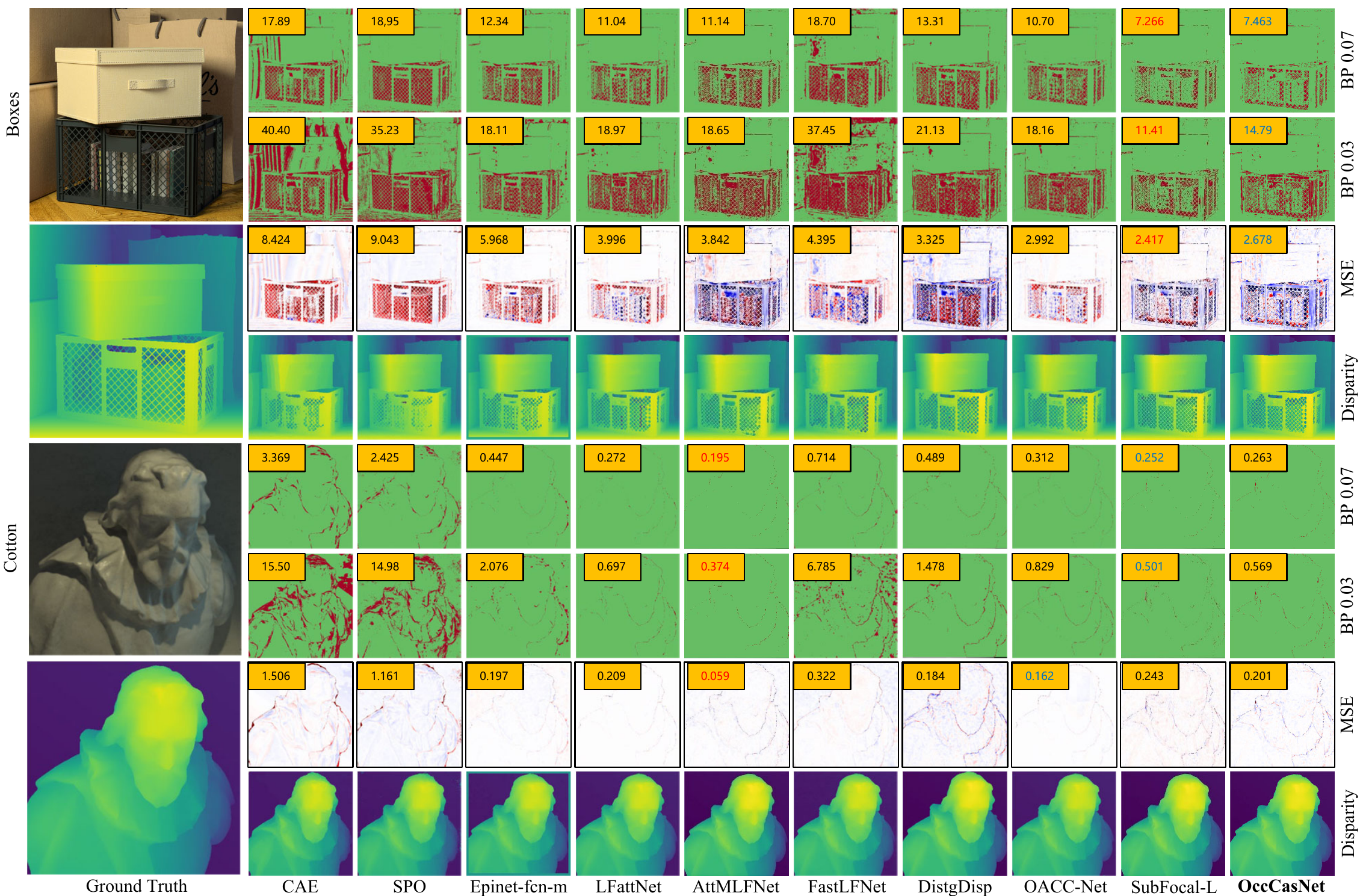}

  \includegraphics[width=0.95\linewidth]{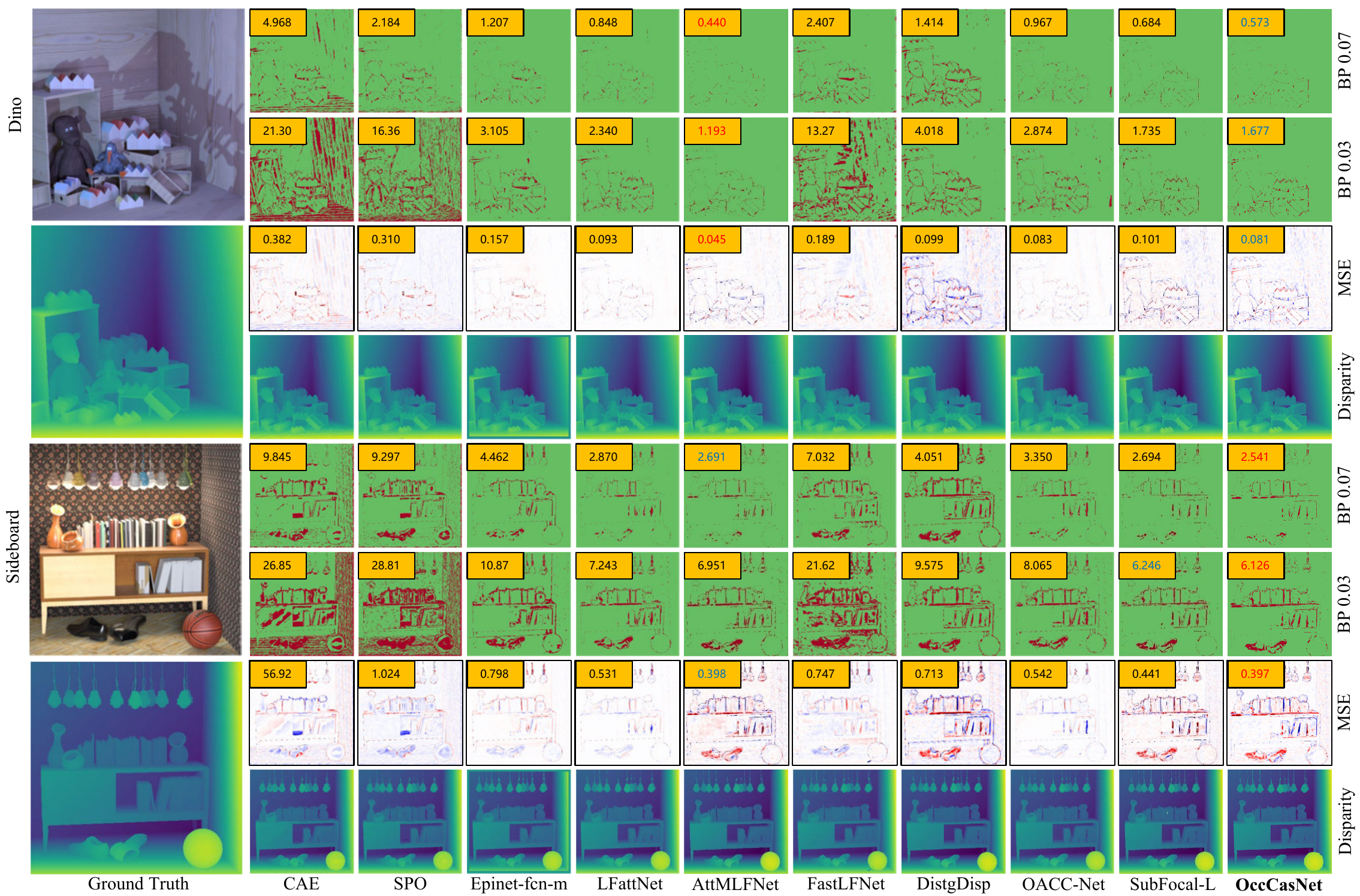}

  \caption{Visual comparisons of disparity and error maps on validation scenes \textit{boxes}, \textit{cotton}, \textit{dino}, and \textit{sideboard} \cite{rerabek2016new}. Corresponding quantitative scores (BadPix0.07, BadPix0.03, and MSE) are reported on the top-left corner of each error map.
  }

  \label{fig: hci_2}
\end{figure*}

\begin{figure*}[htb]
  \centering

  \includegraphics[width=\linewidth]{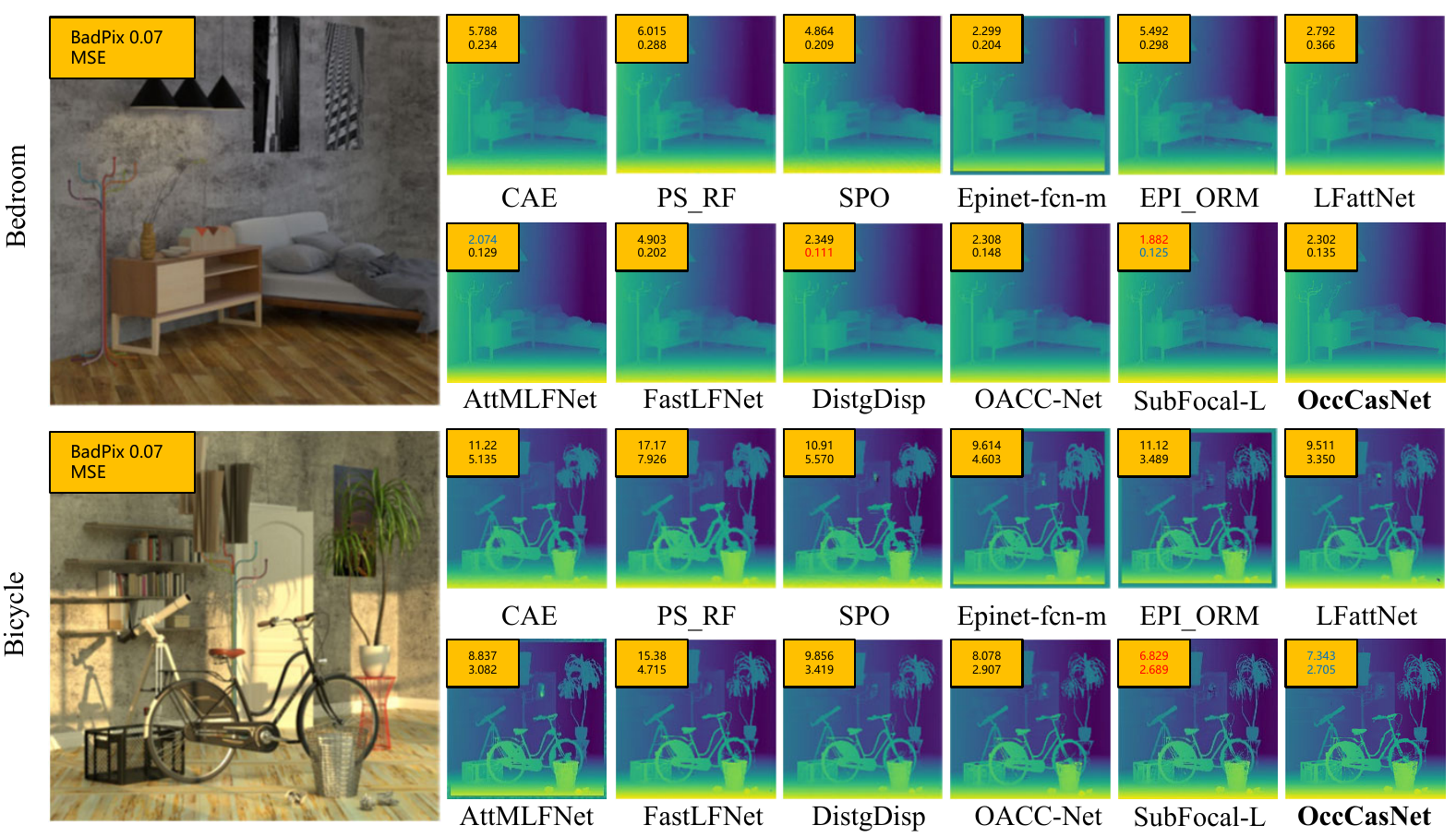}

  \includegraphics[width= \linewidth]{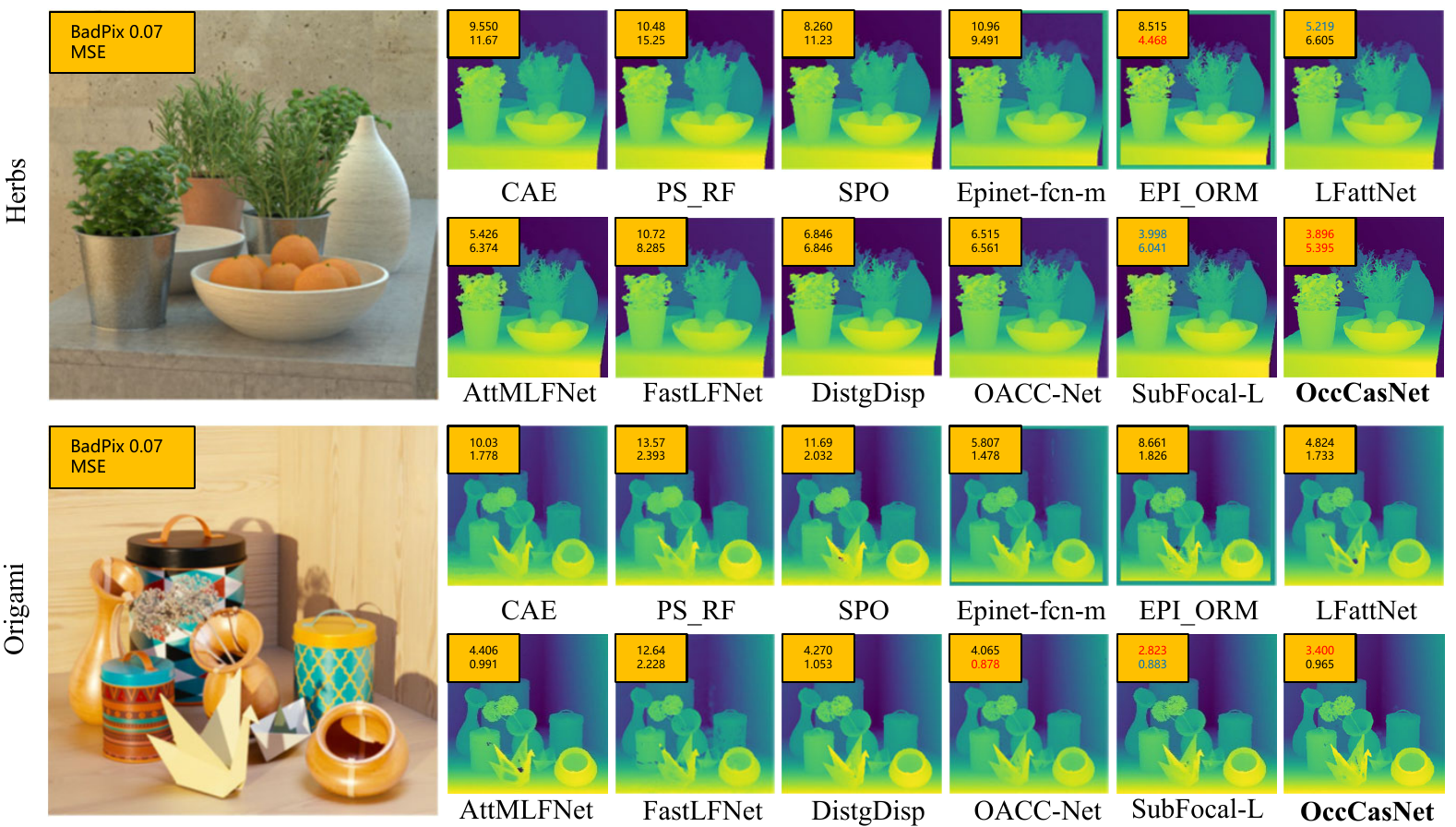}

  \caption{Visual comparisons of disparity maps on test scenes \textit{bedroom}, \textit{bicycle}, \textit{herbs}, and \textit{origami} \cite{rerabek2016new}. The ground-truth disparity maps of these scenes are not released. The BadPix 0.07 and MSE of each method (copied from the benchmark site) are reported on the left-top corner.
  }

  \label{fig: hci_3}
\end{figure*}

Fig. \ref{fig: hci_1} and Fig. \ref{fig: hci_2} depict the estimated disparity maps and error maps for the eight validation scenes. The estimated disparity maps for the four test scenes are shown in Fig. \ref{fig: hci_3}.

\begin{figure*}[htb]
  \centering
  \includegraphics[width=0.9 \linewidth]{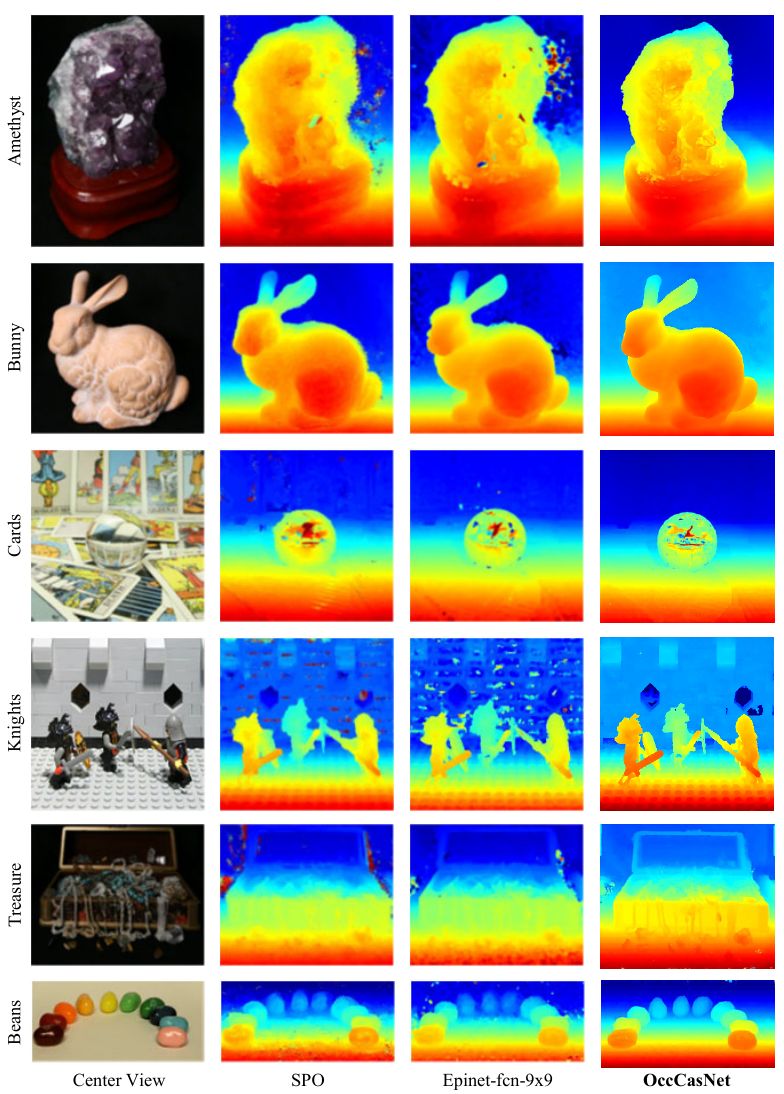}

  \caption{Visual results achieved by SPO\cite{zhang2016robust}, EPINET\cite{shin2018epinet}, and our method on the Stanford Gantry LF dataset \cite{rerabek2016new}. ground-truth disparity maps of these real-world LFs are unavailable.
  }

  \label{fig: real_1}
\end{figure*}

\begin{figure*}[htb]
  \centering
  \includegraphics[width=\linewidth]{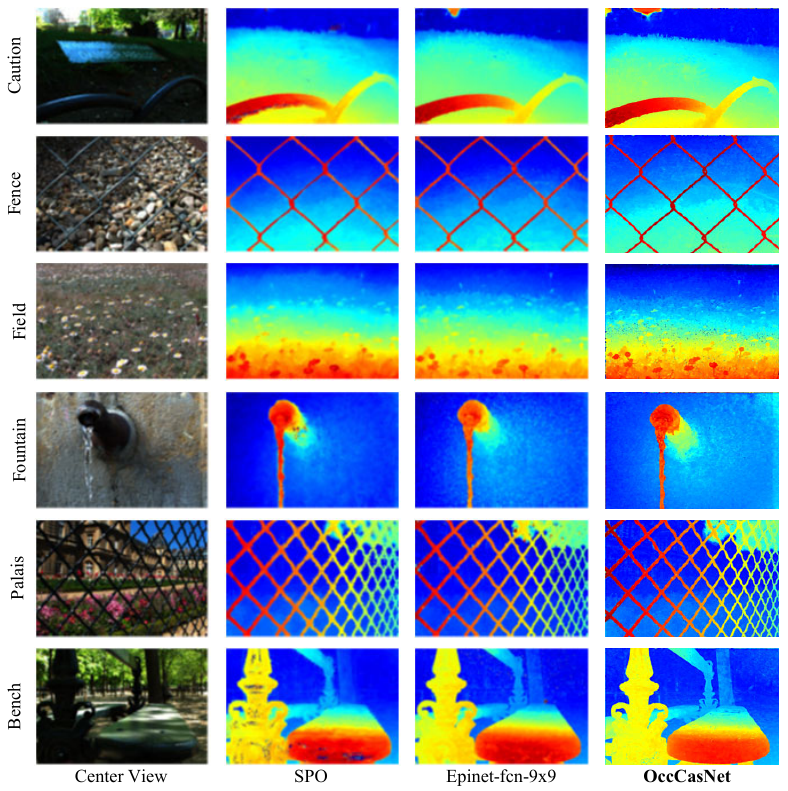}

  \caption{Visual results achieved by SPO \cite{zhang2016robust}, EPINET \cite{shin2018epinet}, and our method on LFs captured by Lytro cameras \cite{le2018light, rerabek2016new}.
  }

  \label{fig: real_2}
\end{figure*}

\begin{figure*}[htb]
  \centering
  \includegraphics[width= 0.88 \linewidth]{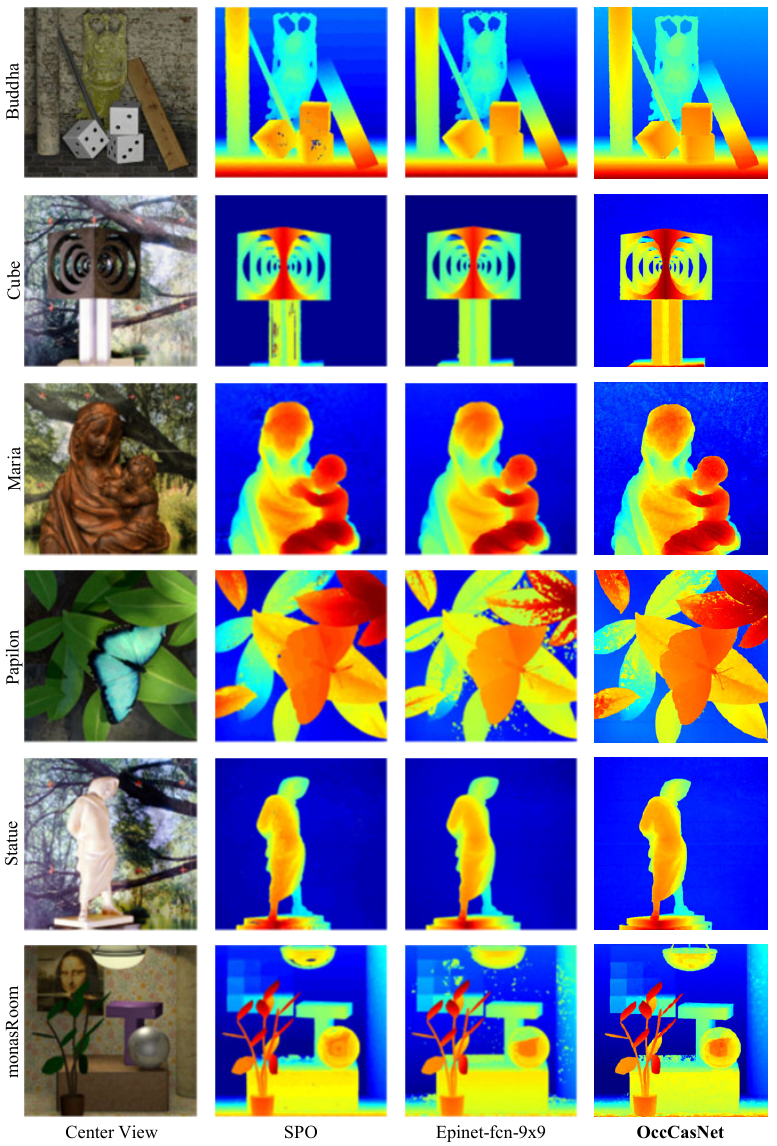}
  \caption{Visual results achieved by SPO \cite{zhang2016robust}, EPINET \cite{shin2018epinet}, 
  and our method on the old HCI LF dataset \cite{vaish2008new}.
  }
  \label{fig: real_3}
\end{figure*}

\subsection{Results on different LF datasets}
\label{sub:visual}

Fig. \ref{fig: real_1}, Fig. \ref{fig: real_2} and Fig. \ref{fig: real_3} compare the visual results obtained by SPO \cite{zhang2016robust}, and EPINET \cite{shin2018epinet} and our method on various types of LF datasets \cite{le2018light,rerabek2016new,vaish2008new,wanner2013datasets}.


 




\end{appendix}

\vfill

\end{document}